\pdfoutput=1

\documentclass{sig-alternate}

\usepackage{times}
\usepackage{caption}
\usepackage{setspace}
\usepackage{enumitem}

\usepackage{subcaption}
\usepackage{hyperref}

\usepackage{diagbox}

\makeatletter
\newcommand\thefontsize[1]{{#1 The current font size is: \f@size pt\par}}
\makeatother

\def\sharedaffiliation{%
\end{tabular}
\begin{tabular}{c}}

\begin{document}

\title{DeepSense: A Unified Deep Learning Framework for Time-Series Mobile Sensing Data Processing}

\numberofauthors{5} 
%
\author{\hspace{0.65cm}
%
%
\alignauthor
Shuochao Yao$^\dagger$\\
\email{syao9@illinois.edu}
\alignauthor
Shaohan Hu$^\ddagger$\\
\email{shaohan.hu@ibm.com}
\alignauthor
Yiran Zhao$^\dagger$\\
\email{zhao97@illinois.edu}
\alignauthor
\and
Aston Zhang$^\dagger$\\
       \email{lzhang74@illinois.edu}
\alignauthor
Tarek Abdelzaher$^\dagger$\\
       \email{zaher@illinois.edu}
\sharedaffiliation
       \affaddr{$^\dagger$University of Illinois at Urbana-Champaign, Urbana, IL USA }   \\
\sharedaffiliation
       \affaddr{$^\ddagger$IBM Research, Yorktown Heights, NY USA}   \\
}
\sloppy
\maketitle

{
\begin{abstract}
Mobile sensing and computing applications usually require time-series inputs from sensors, such as accelerometers, gyroscopes, and magnetometers. Some applications, such as tracking, can use sensed acceleration and rate of rotation to calculate displacement based on physical system models. Other applications, such as activity recognition, extract manually designed features from sensor inputs for classification. Such applications face two challenges. On one hand, on-device sensor measurements are noisy. For many mobile applications, it is hard to find a distribution that exactly describes the noise in practice. Unfortunately, calculating target quantities based on physical system and noise models is only as accurate as the noise assumptions. Similarly, in classification applications, although manually designed features have proven to be effective, it is not always straightforward to find the most robust features to accommodate diverse sensor noise patterns and heterogeneous user behaviors. To this end, we propose DeepSense, a deep learning framework that directly addresses the aforementioned noise and feature customization challenges in a unified manner. DeepSense integrates convolutional and recurrent neural networks to exploit local interactions among similar mobile sensors, merge local interactions of different sensory modalities into global interactions, and extract temporal relationships to model signal dynamics. DeepSense thus provides a general signal estimation and classification framework that accommodates a wide range of applications. We demonstrate the effectiveness of DeepSense using three representative and challenging tasks: car tracking with motion sensors, heterogeneous human activity recognition, and user identification with biometric motion analysis. DeepSense significantly outperforms the state-of-the-art methods for all three tasks. In addition, we show that DeepSense is feasible to implement on smartphones and embedded devices thanks to its moderate energy consumption and low latency.
\end{abstract}

\section{Introduction}
A wide range of mobile sensing and computing applications require time-series measurements from such sensors as accelerometers, gyroscopes, and magnetometers to generate inputs for various signal estimation and classification applications~\cite{lane2010survey}.  Using these sensors, mobile devices are able to infer user activities and states~\cite{fang2016bodyscan, ren2013smartphone, stisen2015smart} and recognize surrounding context~\cite{nath2012ace, xu2013crowd++}. These capabilities serve diverse application areas including health and wellbeing~\cite{ko2010wireless, rabbi2015mybehavior, li2014bioscope}, tracking and imaging~\cite{gowda2016tracking,kjaergaard2011energy,li2015human,zhu2015reusing}, mobile security~\cite{miluzzo2012tapprints, miluzzo2010eyephone, wang2016friend}, and vehicular road sensing~\cite{hu2014towards,hu2015smartroad,kang2015ecodrive}.

Although mobile sensing is becoming increasingly ubiquitous, key challenges remain in improving the accuracy of sensor exploitation. In this paper, we consider the general problem of estimating signals from noisy measurements in mobile sensing applications. This problem can be categorized into two subtypes: regression and classification, depending on whether prediction results are continuous or categorical, respectively.

For regression-oriented problems, such as tracking and localization, sensor inputs are usually processed based on physical models of the phenomena involved. Sensors on mobile devices generate
time-series measurements of physical quantities such as acceleration and angular velocity.
From these measurements, other physical quantities can be computed, such as displacement through double integration of acceleration over time. However, measurements of commodity sensors are noisy. The noise in measurements is non-linear~\cite{ang2007nonlinear} and correlated over time~\cite{park2005error}, which makes it hard to model. This makes it challenging to separate signal from noise, leading to estimation errors and bias.

For classification-oriented problems, such as activity and context recognition, a typical approach is to compute appropriate features derived from raw sensor data. These hand-crafted features are then fed into a classifier for training. This general workflow for classification face the challenge that designing good hand-crafted features can be time consuming; it requires extensive experiments to generalize well to diverse settings such as different sensor noise patterns and heterogeneous user behaviors~\cite{stisen2015smart}.

In this work, we propose DeepSense, a unified deep learning framework that directly addresses the aforementioned customization challenges that arise in mobile sensing applications. The core of  DeepSense is the integration of convolutional neural networks (CNN)  and recurrent neural networks (RNN). Input sensor measurements are split into a series of data intervals along time. The frequency representation of each data intervals is fed into a CNN to learn intra-interval local interactions within each sensing modality and intra-interval global interactions among different sensor inputs, hierarchically. The intra-interval representations along time are then fed into an RNN to learn the inter-interval relationships. The whole framework can be easily customized to fit specific mobile computing (regression or classification) tasks by three simple steps, as will be described later.

For the regression-oriented mobile sensing problem, DeepSense learns the composition of physical system and noise model to yield outputs from noisy sensor data directly. The neural network acts as an approximate transfer function. The CNN part approximates the computation of sensing quantities within the time interval, and the RNN part approximates the computation of sensing quantities across time intervals.

For the classification-oriented mobile sensing problem, the neural network acts as an automatic feature extractor encoding local, global, and temporal information.
The CNN part extracts local features within each sensor modality and merges the local features of different sensory modalities into global features hierarchically. The RNN part extracts temporal dependencies.

We demonstrate the effectiveness of our DeepSense framework using three representative and challenging mobile sensing problems, which illustrate the potential of solving different tasks with a single unified modeling methodology:
\begin{itemize}
\vspace{-0.1cm}
\item {\em Car tracking with motion sensors:\/} In this task, we use dead reckoning to infer position from acceleration measurements. One of the major contributions of DeepSense is its ability to withstand nonlinear and time-dependent noise and bias. We chose the car tracking task because it involves double-integration and thus is particularly sensitive to error accumulation, as acceleration errors can lead to significant deviations in position estimate over time. This task thus constitutes a worst-case of sorts in terms of emphasizing the effects of noise on modelling error. Traditionally, external means are needed to reset the error when possible~\cite{chandrasekaran2011tracking, hu2015experiences, lin2010energy}. We intentionally forgo such means to demostrate the capability of DeepSense for learning accurate models of target quantities in the presence of realistic noise.
\vspace{-0.1cm}
\item {\em Heterogeneous human activity recognition:\/}
Although human activity recognition with motion sensors is a mature problem, Allan {\em et al.\/}~\cite{stisen2015smart} illustrated that state-of-the-art algorithms do not generalize well across users when a new user is tested who has not appeared in the training set. This classification-oriented problem therefore illustrates the capability of DeepSense to extract features that generalize better across users in mobile sensing tasks.
\vspace{-0.1cm}
\item {\em User identification with biometric motion analysis:\/}
Biometric gait analysis can be used to identify users when they are walking~\cite{gadaleta2016idnet,ren2013smartphone}. We extend walking to other activities, such as  biking and climbing stairs, for user identification. This classification-oriented problem illustrates the capability of DeepSense to extract distinct features for different users or classes.
\end{itemize}

We evaluate these three tasks with collected data or existing datasets. We compare DeepSense to state-of-the-art algorithms that solve the respective tasks, as well as to three DeepSense variants, each presenting a simplification of the algorithm as described in Section~\ref{sec:baselines}.
For the regression-oriented problem: car tracking with motion sensors, DeepSense provides an estimator with far smaller tracking error. This makes tracking with solely noisy on-device motion sensors practical and illustrates the capability of DeepSense to perform accurate estimation of physical quantities from noisy sensor data. For the other two classification-oriented problems, DeepSense outperforms state-of-the-art algorithms by a large margin, illustrating its capability to automatically learn robust and distinct features. DeepSense outperforms all its simpler variants in all three tasks, which shows the effectiveness of its design components.
Despite a general shift towards remote cloud processing for a range of mobile applications, we argue that it is intrinsically desirable that heavy sensing tasks be carried out locally on-device, due to the usually tight latency requirements, and the prohibitively large data transmission requirement as dictated by the high sensor sampling frequency (e.g. accelerometer, gyroscope).
Therefore, we also demonstrate the feasibility of implementing and deploying
DeepSense on mobile devices by showing its moderate energy consumption and low overhead for all three tasks on two different types of smart devices.

In summary, the main contribution of this paper is that {\em we develop a deep learning framework, DeepSense, that solves both regression-oriented and classification-oriented mobile computing tasks in a unified manner. By exploiting local interactions within each sensing modality, merging local interactions of different sensing modalities into global interactions, and extracting temporal relationships,
DeepSense learns the composition of physical laws and noise model in regression-oriented problems, and automatically extracts robust and distinct features that contain local, global, and temporal relationships in classification-oriented problems. Importantly, it outperforms the state of the art, while remaining implementable on mobile devices.\/}

The rest of this paper is organized as follows. Section~\ref{sec:relatedWork} introduces related work on deep learning in the context of mobile sensing and computing. We describe the technical details of DeepSense in Section~\ref{sec:model} and the way to customize DeepSense to mobile computing problems in Section~\ref{sec:indvArch}. The evaluation is presented in Section~\ref{sec:evaluation}. Finally, we discuss the results in Section~\ref{sec:discussion} and conclude in Section~\ref{sec:conclusion}.

\section{Related Work}~\label{sec:relatedWork}
Recently, deep learning~\cite{Goodfellow-et-al-2016-Book} has become one of the most popular methodologies in AI-related tasks, such as computer vision~\cite{he2015deep}, speech recognition~\cite{dahl2012context}, and natural language processing~\cite{bahdanau2014neural}. Lots of deep learning architectures have been proposed to exploit the relationships embedded in different types of inputs. For example, Residual nets~\cite{he2015deep} introduce shortcut connections into CNNs, which greatly reduces the difficulty of training super-deep models. However, since residual nets mainly focus on visual inputs, they lose the capability to model temporal relationships, which are of great importance in time-series sensor inputs. LRCNs~\cite{donahue2015long} apply CNNs to extract features for each video frame and combine video frame sequences with LSTM~\cite{greff2015lstm}, which exploits spatio-temporal relationships in video inputs. However, it does not consider modeling multimodal inputs. This capability is important to mobile sensing and computing tasks, because most tasks require collaboration among multiple sensors.  Multimodal DBMs~\cite{srivastava2012multimodal} merge multimodal inputs, such as images and text, with Deep Boltzmann Machines (DBMs). However, the work does not model temporal relationships and does not apply tailored structures, such as CNNs, to effectively and efficiently exploit local interactions within input data. To the best of our knowledge, DeepSense is the first architecture that possesses the capability for both (i) modelling temporal relationships and (ii) fusing multimodal sensor inputs. It also contains specifically designed structures to exploit local interactions in sensor inputs.

\begin{figure*}[!htb]
\vspace{-0.1cm}
\centering
\includegraphics[width=0.95\linewidth]{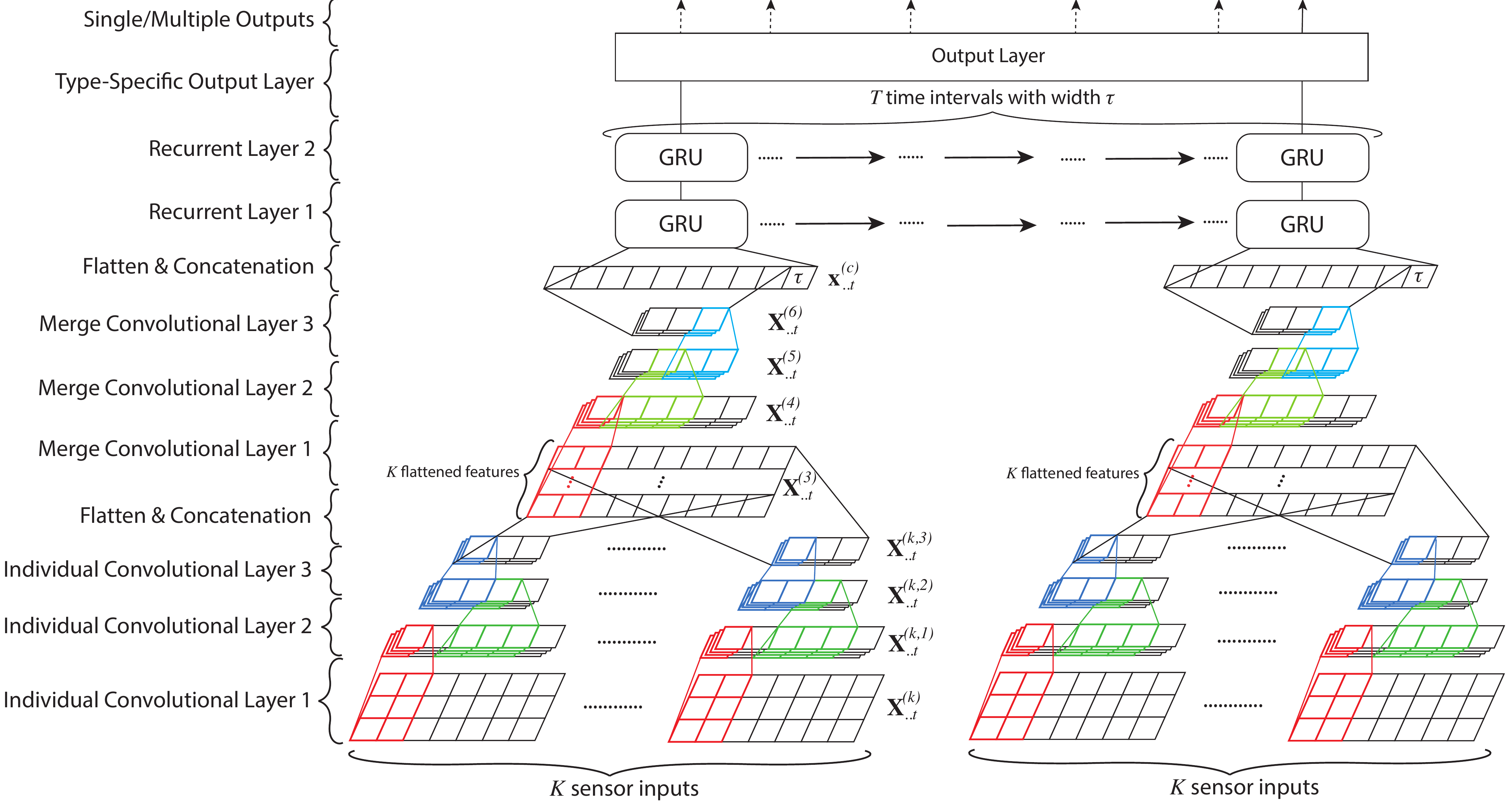}
\caption{Main architecture of the DeepSense framework.}
\label{fig:mainArch}
\vspace{-0.3cm}
\end{figure*}

There are several illuminating studies, applying deep neural network models to different mobile sensing applications. DeepEar~\cite{lane2015deepear} uses Deep Boltzmann Machines to improve the performance of audio sensing tasks in an environment with background noise. RBM~\cite{bhattacharya2016smart} and MultiRBM~\cite{radu2016towards} use Deep Boltzmann Machines and Multimodal DBMs to improve the performance of heterogeneous human activity recognition. IDNet~\cite{gadaleta2016idnet} applies CNNs to the biometric gait analysis task. DeepX~\cite{lane2016deepx} and RedEye~\cite{likamwa2016redeye} reduce the energy consumption of deep neural networks, based on software and hardware, respectively. However, these studies do not capture the temporal relationships in time-series sensor inputs, and, with the only exception of MultiRBM, lack the capability of fusing multimodal sensor inputs. In addition, these techniques focus on classification-oriented tasks only. To the best of our knowledge, DeepSense is the first framework that directly solves both regression-based and classification-based problems in a unified manner.

\section{DeepSense Framework}~\label{sec:model}
We introduce DeepSense, a unified framework for mobile applications with sensor data inputs, in this section.
We separate our description into three parts. The first two parts, convolutional layers and recurrent layers, are the main building blocks for DeepSense, which are the same for all applications. The third part, the output layer, is the specific layer for two different types of applications; regression-oriented and classification-oriented.

For the rest of this paper, all vectors are denoted by bold lower-case letters (e.g., $\mathbf{x}$ and $\mathbf{y}$), while matrices and tensors are represented by bold upper-case letters (e.g., $\mathbf{X}$ and $\mathbf{Y}$). For a vector $\mathbf{x}$, the $j^{th}$ element is denoted by $\mathbf{x}_{[j]}$. For a tensor $\mathbf{X}$, the $t^{th}$ matrix along the third axis is denoted by $\mathbf{X}_{\cdot \cdot t}$, and other slicing denotations are defined similarly. We use calligraphic letters to denote sets (e.g., $\mathcal{X}$ and $\mathcal{Y}$). For any set $\mathcal{X}$, $|\mathcal{X}|$ denotes the cardinality of $\mathcal{X}$.

For a particular application, we assume that there are $K$ different types of input sensors $\mathcal{S} = \{S_k\}$, $k \in \{1, \cdots, K\}$. Take a sensor $S_k$ as an example. It generates a series of measurements over time. The measurements can be represented by a $d^{(k)} \times n^{(k)}$ matrix $\mathbf{V}$ for measured values and $n^{(k)}$-dimensional vector $\mathbf{u}$ for time stamps, where $d^{(k)}$ is the dimension for each measurement (e.g., measurements along x, y, and z axes for motion sensors) and $n^{(k)}$ is the number of measurements. We split the input measurements $\mathbf{V}$ and $\mathbf{u}$ along time (i.e., columns for $\mathbf{V}$) to generate a series of non-overlapping time intervals with width $\tau$, $\mathcal{W} = \{(\mathbf{V}_t^{(k)}, \mathbf{u}_t^{(k)})\}$, where $|\mathcal{W}| = T$ . Note that, $\tau$ can be different for different intervals, but here we assume a fixed time interval width for succinctness. We then apply Fourier transform to each element in $\mathcal{W}$, because the frequency domain contains better local frequency patterns that are independent of how time-series data is organized in the time domain. We stack these outputs into a $d^{(k)}\times 2f \times T$ tensor $\mathbf{X}^{(k)}$, where $f$ is the dimension of frequency domain containing $f$ magnitude and phase pairs. The set of resulting tensors for each sensor, $\mathcal{X} = \{\mathbf{X}^{(k)}\}$, is the input of DeepSense.

As shown in Fig.~\ref{fig:mainArch}, DeepSense has three major components; the convolutional layers, the recurrent layers, and the output layer, stacked from bottom to top. In the following subsections, we detail these components, respectively.

\subsection{Convolutional Layers}

The convolutional layers can be further separated into two parts: an individual convolutional subnet for each input sensor tensor $\mathbf{X}^{(k)}$, and a single merge convolutional subnet for the output of $K$ individual convolutional subnets' outputs.

Since the structures of individual convolutional subnet for different sensors are the same, we focus on one individual convolutional subnet with input tensor $\mathbf{X}^{(k)}$. Recall that $\mathbf{X}^{(k)}$ is a $d^{(k)}\times 2f \times T$ tensor, where $d^{(k)}$ is the sensor measurement dimension, $f$ is the dimension of frequency domain, and $T$ is the number of time intervals. For each time interval $t$, the matrix $\mathbf{X}^{(k)}_{\cdot\cdot t}$ will be fed into a CNN architecture (with three layers in this paper). There are two kinds of features/relationships embedded in $\mathbf{X}^{(k)}_{\cdot\cdot t}$ we want to extract. The relationships within the frequency domain and across sensor measurement dimension. The frequency domain usually contains lots of local patterns in some neighbouring frequencies. And the interaction among sensor measurement usually including all dimensions. Therefore, we first apply 2d filters with shape $(d^{(k)}, cov1)$ to $\mathbf{X}^{(k)}_{\cdot\cdot t}$ to learn interaction among sensor measurement dimensions and local patterns in frequency domain, with the output $\mathbf{X}^{(k,1)}_{\cdot\cdot t}$. Then we apply 1d filters with shape $(1, cov2)$ and $(1, cov3)$ hierarchically to learn high-level relationships, $\mathbf{X}^{(k,2)}_{\cdot\cdot t}$ and $\mathbf{X}^{(k,3)}_{\cdot\cdot t}$.

Then we flatten matrix $\mathbf{X}^{(k,3)}_{\cdot\cdot t}$ into vector $\mathbf{x}^{(k,3)}_{\cdot\cdot t}$ and concat all $K$ vectors $\{\mathbf{x}^{(k,3)}_{\cdot\cdot t}\}$ into a $K$-row matrix $\mathbf{X}^{(3)}_{\cdot\cdot t}$, which is the input of the merge convolutional subnet. The architecture of the merge convolutional subnet is similar as the individual convolutional subnet. We first apply 2d filters with shape $(K, cov4)$ to learn the interactions among all $K$ sensors, with output $\mathbf{X}^{(4)}_{\cdot\cdot t}$, and then apply 1d filters with shape $(1, cov5)$ and $(1, cov6)$ hierarchically to learn high-level relationships, $\mathbf{X}^{(5)}_{\cdot\cdot t}$ and $\mathbf{X}^{(6)}_{\cdot\cdot t}$.

For each convolutional layer, DeepSense learns $64$ filters, and uses ReLU as the activation function. In addition, batch normalization~\cite{ioffe2015batch} is applied at each layer to reduce internal covariate shift. We do not use residual net structures~\cite{he2015deep}, because we want to simplify the network architecture for mobile applications. Then we flatten the final output $\mathbf{X}^{(6)}_{\cdot\cdot t}$ into vector $\mathbf{x}^{(f)}_{\cdot\cdot t}$; concatenate $\mathbf{x}^{(f)}_{\cdot\cdot t}$ and time interval width, $[\tau]$, together into $\mathbf{x}^{(c)}_t$ as inputs of recurrent layers.

\subsection{Recurrent Layers}
Recurrent neural networks are powerful architectures that can approximate function and learn meaningful features for sequences. Original RNNs fall short of learning long-term dependencies. Two extended models are  Long Short-Term Memory (LSTM)~\cite{greff2015lstm} and Gated Recurrent Unit (GRU)~\cite{chung2014empirical}. In this paper, we choose GRU, because GRUs show similar performance as LSTMs on various tasks~\cite{chung2014empirical}, while having a more concise expression, which reduces network complexity for mobile applications.

DeepSense chooses a stacked GRU structure (with two layers in this paper). Compared with standard (single-layer) GRUs, stacked GRUs are a more efficient way to increase model capacity~\cite{Goodfellow-et-al-2016-Book}. Compared to bidirectional GRUs~\cite{schuster1997bidirectional}, which contain two time flows from start to end and from end to start, stacked GRUs can run incrementally, when there is a new time interval, resulting in faster processing of stream data. In contrast, we cannot run bidirectional GRUs until data from all time intervals are ready, which is infeasible for applications such as tracking. We apply dropout to the connections between GRU layers~\cite{zaremba2014recurrent} for regularization and apply recurrent batch normalization~\cite{cooijmans2016recurrent} to reduce internal covariate shift among time steps. Inputs $\{\mathbf{x}_t^{(c)}\}$ for $t = 1, \cdots, T$ from previous convolutional layers are fed into stacked GRU and  generate outputs $\{\mathbf{x}_t^{(r)}\}$ for $t = 1, \cdots, T$ as inputs of the final output layer.

\subsection{Output Layer}
The output of recurrent layer is a series of vectors $\{\mathbf{x}_t^{(r)}\}$ for $t = 1, \cdots, T$.
For the regression-oriented task, since the value of each element in vector $\mathbf{x}_t^{(r)}$ is within $\pm1$, $\mathbf{x}_t^{(r)}$ encodes the output physical quantities at the end of time interval $t$. In the output layer, we want to learn a dictionary $\mathbf{W}_{out}$ with a bias term $\mathbf{b}_{out}$ to decode $\mathbf{x}_t^{(r)}$  into $\mathbf{\hat{y}}_t$, such that $\mathbf{\hat{y}}_t = \mathbf{W}_{out} \cdot \mathbf{x}_t^{(r)} + \mathbf{b}_{out}$. Therefore, the output layer is a fully connected layer on the top of each interval with sharing parameter $\mathbf{W}_{out}$  and $\mathbf{b}_{out}$.

For the classification task, $\mathbf{x}_t^{(r)}$ is the feature vector at time interval $t$. The output layer first needs to compose $\{\mathbf{x}_t^{(r)}\}$ into a fixed-length feature vector for further processing. Averaging features over time is one choice. More sophisticated methods can also be applied to generate the final feature, such as the attention model~\cite{bahdanau2014neural}, which has illustrated its effectiveness in various learning tasks recently. The attention model can be viewed as weighted averaging of features over time, but the weights are learnt by neural networks through context.
In this paper, we still use averaging features over time to generate the final feature, $\mathbf{x}^{(r)} =  (\sum_{t=1}^T \mathbf{x}_t^{(r)})/T$. Then we feed $\mathbf{x}^{(r)}$ into a softmax layer to generate the predicted category probability $\mathbf{\hat{y}}$.

\section{Task-Specific Customization}\label{sec:indvArch}
In this section, we first describe how to trivially customize the DeepSense framework to different mobile sensing and computing tasks. Next, we instantiate the solution with three specific tasks used in our evaluation.

\subsection{General Customization Process}
In general, we need to customize a few parameters of the main architecture of DeepSense, shown in Section~\ref{sec:model}, for specific mobile sensing and computing tasks. Our general DeepSense customization process is as follows:
\begin{enumerate}[leftmargin=*]
\item Identify the number of sensor inputs, $K$. Pre-process the sensor inputs into a set of tensors $\mathcal{X} = \{\mathbf{X}^{(k)}\}$ as input.
\item Identify the type of the task. Whether the application is regression or classification-oriented. Select one of the two types of output layer according to the type of task.
\item Design a customized cost function or choose the default cost function (namely, mean square error for regression-oriented tasks and cross-entropy error for classification-oriented tasks).
\end{enumerate}

Therefore, if opt for the default DeepSense configuration, we need only to set the number of inputs, $K$, preprocess the input sensor measurements, and identify the type of task (i.e., regression-oriented versus classification-oriented).

The pre-processing is simple, as stated at the beginning of Section~\ref{sec:model}.
We just need to align and chunk the sensor measurements, and apply Fourier transform to each sensor chunk. For each sensor, we stack these frequency domain outputs into $d^{(k)}\times 2f \times T$ tensor $\mathbf{X}^{(k)}$, where $d^{(k)}$ is the sensor measurement dimension, $f$ is the frequency domain dimension, and $T$ is the number of time intervals.

To identify the number of sensor inputs $K$, we usually set $K$ to be the number of different sensing modalities available. If there exist two or more sensors of the same modality (e.g., two accelerometers or three microphones), we just treat them as one multi-dimensional sensor and set its measurement dimension accordingly.

For the cost function, we can design our own cost function other than the default one. We denote our DeepSense model as function $\mathcal{F}(\cdot)$, and a single training sample pair as $(\mathcal{X}, \mathbf{y})$. We can express the cost function as:
\vspace{-0.2cm}
\begin{eqnarray}
\mathcal{L}  = \ell(\mathcal{F}(\mathcal{X}), \mathbf{y}) + \sum_{j} \lambda_j P_j \label{eqn:lossFunction}
\end{eqnarray}
\vspace{-0.4cm}

\noindent
where $\ell(\cdot)$ is the loss function, $P_j$ is the penalty or regularization function, and $\lambda_j$ controls the importance of the penalty or regularization term.

\subsection{Customize Mobile Sensing Tasks}\label{sec:mobileTaskTrain}
In this section, we provide three instances of customizing DeepSense for specific mobile computing applications used in our evaluation.

\noindent
\textbf{Car tracking with motion sensors (CarTrack): } In this task, we apply accelerator, gyroscope, and magnetometer to track the trajectory of a car without initial speed. Therefore, according to our general customization process, carTrack is a regression-oriented problem with $K=3$ (i.e. accelerometer, gyroscope, and magnetometer). Instead of applying default mean square error loss function, we design our own cost function according to Equation~(\ref{eqn:lossFunction}).

During the training step,  the ground-truth 2D displacement of car in each time interval, $\mathbf{y}$, is obtained by GPS signal, where $\mathbf{y}_{[t]}$ denotes the 2D displacement in time interval $t$. Yet a problem is that GPS signal also contains noise. Training the DeepSense model to recover the displacement obtained from by GPS signal will generate sub-optimal results. We apply Kalman filter to covert displacement $\mathbf{y}_{[t]}$ into a 2D Gaussian distribution $\mathbf{Y}_{[t]}(\cdot)$
with mean value $\mathbf{y}^{(t)}$
in time interval $t$. Therefore, we use negative log likelihood as loss function $\ell(\cdot)$ with additional penalty terms:
\vspace{-0.15cm}
\begin{eqnarray}
\mathcal{L} & =  & -\log\big(\mathbf{Y}_{[t]}\big(\mathcal{F}(\mathcal{X})_{[t]}\big)\big) \notag\\[-0.15cm]
&&+\sum_{t=1}^T \lambda \cdot \max\big(0, \cos(\theta) - S_c\big(\mathcal{F}(\mathcal{X})_{[t]}, \mathbf{y}^{(t)}\big)\big) \notag
\end{eqnarray}
\vspace{-0.3cm}

\noindent
where $S_c(\cdot, \cdot)$ denotes the cosine similarity, the first term is the negative log likelihood loss function, and the second term is a penalty term controlled by parameter $\lambda$. If the angle between our predicted displacement $\mathcal{F}(\mathcal{X})_{[t]}$ and $\mathbf{y}^{(t)}$ is larger than a pre-defined margin $\theta \in [0, \pi)$, the cost function will get a penalty. We introduce the penalty, because we find that predicting a correct direction is more important
during the experiment, as described in Section~\ref{sec:carTrackAcc}.

\noindent
\textbf{Heterogeneous Human activity recognition (HHAR): } In this task, we perform leave-one-user-out cross-validation on human activity recognition task with accelerometer and gyroscope measurements. Therefore, according to our general customization process, HHAR is a classification-oriented problem with $K=2$ (accelerometer and gyroscope). We use the default cross-entropy cost function as the training objective.
\vspace{-0.15cm}
\begin{eqnarray}
\mathcal{L} = H(\mathbf{y}, \mathcal{F}(\mathcal{X})) \notag
\end{eqnarray}
\vspace{-0.5cm}

\noindent
where $H(\cdot, \cdot)$ is the cross entropy for two distributions.

\noindent
\textbf{User Identification with motion analysis (UserID): }In this task, we perform user identification with biometric motion analysis. We classify users' identity according to accelerometer and gyroscope measurements. Similarly, according to our general customization process, UserID is a classification-oriented problem with $K=2$ (accelerometer and gyroscope). Similarly as above, we use the default cross-entropy cost function as the training objective.

\section{Evaluation}~\label{sec:evaluation}
In this section, we evaluate DeepSense on three mobile computing tasks. We first introduce the experimental setup for each, including datasets and baseline algorithms. We then evaluate the three tasks based on accuracy, energy, and latency. We use the abbreviations, CarTrack, HHAR, and UserID, as introduced in Section~\ref{sec:mobileTaskTrain}, to refer to the aforementioned tasks.

\subsection{Data Collection and Datasets}\label{sec:datasets}
For the CarTrack task, we collect 17,500 phone-miles worth of driving data.
Namely, we collect around 500 driving hours in total using three cars fitted with 20 mobile phones in the Urbana-Champaign area. Mobile devices include Nexus 5, Nexus 4, Galaxy Nexus, and Nexus S. Each mobile device collects measures of accelerometer, gyroscope, magnetometer, and GPS. GPS measurements are collected roughly every second. Collection rates of other sensors are set to their highest frequency. After obtaining the raw sensor measurements, we first segment them into data samples. Each data sample is a zero-speed to zero-speed journey, where the start and termination are detected when  there are at least three consecutive zero GPS speed readings. Each data sample is then separated into time intervals according to the GPS measurements. Hence, every GPS measurement is an indicator of the end of a time interval.
In addition, each data sample contains one additional time interval with zero speed at the beginning. Furthermore, for each time interval, GPS latitude and longitude are converted into map coordinates, where the origin of coordinates is the position at the first time interval. Fourier transform is applied to each sensor measurement in each time interval to obtain the frequency response of the three sensing axes. The frequency responses of the accelerator, gyroscope, and magnetometer at each time interval are then composed into the tensors as DeepSense inputs. At last, for evaluation purposes, we apply a Kalman filter to coordinates obtained by the GPS signal, and generate the displacement distribution of each time interval. The results serve as ground truth for training.

For both the HHAR and UserID tasks, we use the dataset collected by Allan et al.~\cite{stisen2015smart}. This dataset contains readings from two motion sensors (accelerometer and gyroscope). Readings were recorded when users executed activities scripted in no specific order, while carrying smartwatches and smartphones. The dataset contains 9 users, 6 activities (biking, sitting, standing, walking, climbStair-up, and climbStair-down), and 6 types of mobile devices. For both tasks, accelerometer and gyroscope measurements are model inputs. However, for HHAR, activities are used as labels, and for UserID, users' unique IDs are used as labels.
We segment raw measurements into 5-second samples.
For DeepSense, each sample is further divided into time intervals of length $\tau$, as shown in Figure~\ref{fig:mainArch}. We take $\tau = 0.25$ s.
Then we calculate the frequency response of sensors for each time interval, and compose results from different time intervals into tensors as inputs.

\subsection{Algorithms in Comparison}\label{sec:baselines}
We evaluate our DeepSense model and compare it with other competitive algorithms in three tasks. There are three global baselines, which are the variants of DeepSense model by removing one design component in the architecture. The other baselines are specifically designed for each single task.

\noindent
\textbf{DS-singleGRU:}
This model replaces the 2-layer stacked GRU with a single-layer GRU with larger dimension, while keeping the number of parameters. This baseline algorithm is used to verify the efficiency of increasing model capacity by staked recurrent layer.

\noindent
\textbf{DS-noIndvConv:}
In this mode, there are no individual convolutional subnets for each sensor input. Instead, we concatenate the input tensors along the first axis (i.e., the input measurement dimension). Then, for each time interval, we have a single matrix as the input to the merge convolutional subnet directly.

\noindent
\textbf{DS-noMergeConv:}
In this variant, there are no merge convolutional subnets at each time interval. Instead, we flatten the output of each individual convolutional subnet and concatenate them into a single vector as the input of the recurrent layers.

\noindent
\textbf{CarTrack Baseline:}

$\bullet$ \textbf{GPS:} This is a baseline measurement that is specific to the CarTrack problem. It can be viewed as the ground truth for the task, as we do not have other means of more accurately acquiring cars' locations.

$\bullet$ \textbf{Sensor-fusion:} This is a sensor fusion based algorithm. It combines gyroscope and accelerometer measurements to obtain the pure acceleration without gravity. It uses accelerometer, gyroscope, and magnetometer to obtain absolute rotation calibration. Android phones have proprietary solutions for these two functions~\cite{milette2012professional}. The algorithm then applies double integration on pure acceleration with absolute rotation calibration to obtain the displacement.

$\bullet$ \textbf{eNav (w/o GPS):} eNav is a map-aided car tracking algorithm~\cite{hu2015experiences}. This algorithm constrains the car movement path according to a digital map, and computes moving distance along the path using double integration of acceleration derived using principal component analysis that removes gravity. The original eNav uses GPS when it believes that dead-reckoning error is high. For fairness, we modified eNav to disable GPS.

\noindent
\textbf{HHAR Baselines:}

$\bullet$ \textbf{HAR-RF:}
This algorithm~\cite{stisen2015smart} selects all popular time-domain and frequency domain features from~\cite{figo2010preprocessing} and ECDF features from~\cite{hammerla2013preserving}, and uses random forest as classifier.

$\bullet$ \textbf{HAR-SVM:}
Feature selection of this model is same as the HAR-RF model. But this model uses support vector machine as classifier~\cite{stisen2015smart}.

$\bullet$ \textbf{HRA-RBM:}
This model is based on stacked restricted Boltzmann machines with frequency domain representations as inputs~\cite{bhattacharya2016smart}.

$\bullet$ \textbf{HRA-MultiRBM:}
For each sensor input, the model processes it with a single stacked restricted Boltzmann machine. Then it uses another stacked restricted Boltzmann machine to merge the results for activity recognition~\cite{radu2016towards}.

\noindent
\textbf{UserID Baselines:}

$\bullet$ \textbf{GaitID:}
This model extracts the gait template and identifies user through template matching with support vector machine~\cite{thang2012gait}.

$\bullet$ \textbf{IDNet:}
This model first extracts the gait template, and extracts template features with convolutional neural networks. Then this model identifies user through support vector machine and integrates multiple verifications with Wald's probability ratio test~\cite{gadaleta2016idnet}.

\subsection{Effectiveness}
In this section, we will discuss the accuracy and other related performance metrics of the DeepSense model, compared with other baseline algorithms.

\begin{figure}[!htb]
\vspace{-0.2cm}
\begin{minipage}[b]{40mm}
\centering
\includegraphics[width=1.\linewidth]{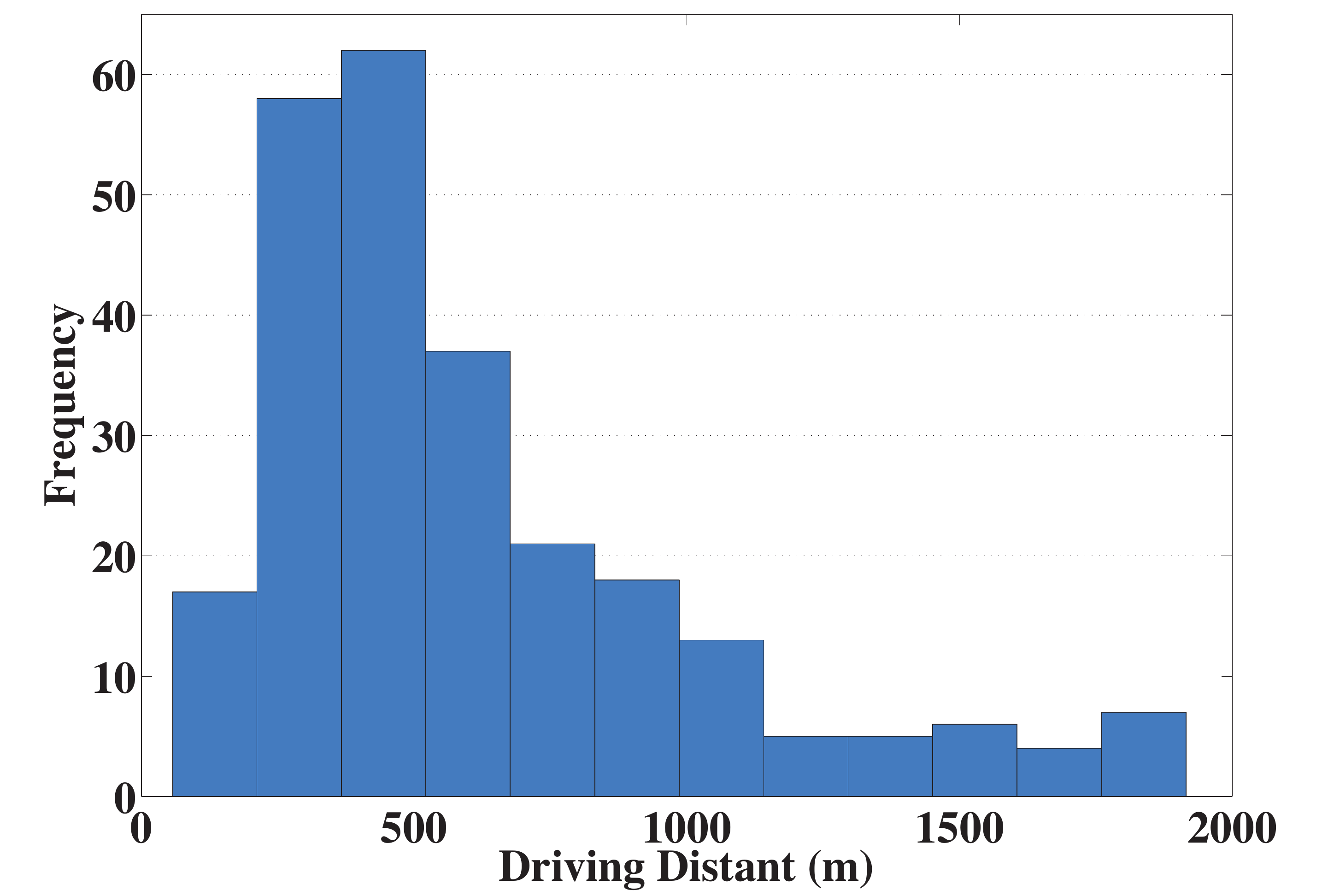}
\caption{Histogram of Driving Distance.}
\label{fig:carHistgram}
\end{minipage}
\begin{minipage}[b]{40mm}
\centering
\includegraphics[width=1.\linewidth]{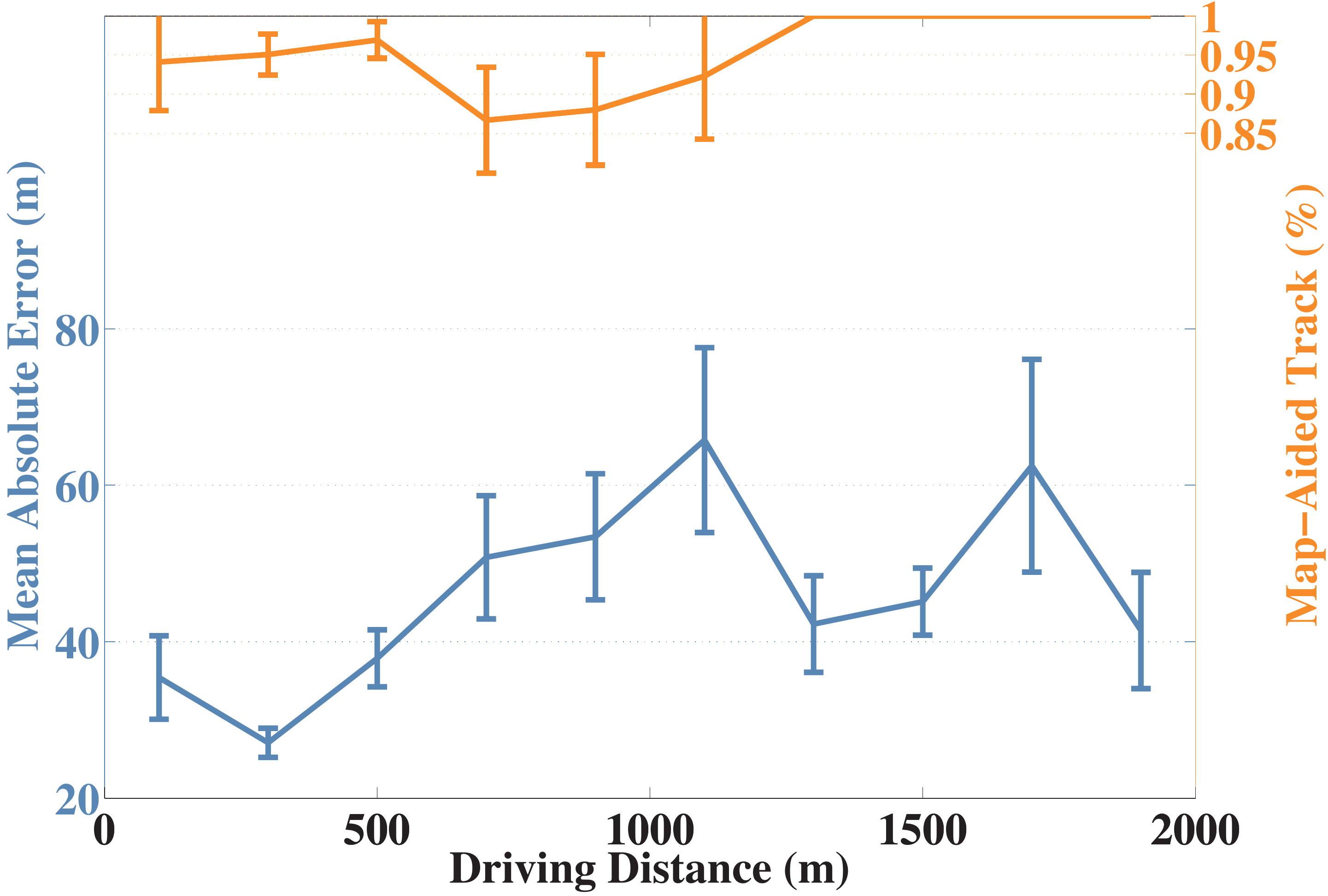}
\caption{Performance over driving distance.}
\label{fig:carTrack-metricDistant}
\end{minipage}
\vspace{-0.3cm}
\end{figure}

\subsubsection{CarTrack}~\label{sec:carTrackAcc}
We use 253 zero-speed to zero-speed car driving examples to evaluate the CarTrack task. The histogram of evaluation data driving distance is illustrated in Fig.~\ref{fig:carHistgram}.

\begin{table}[!htb]
\vspace{-0.1cm}
\centering
\caption{CarTrack Task Accuracy}\label{tab:carMAE}
\vspace{-0.15cm}
\begin{tabular}{|c || c | c| }
\hline
 & MAE (meter) & Map-Aided Accuracy \\
\hline
DeepSense & $\mathbf{40.43\pm 5.24}$ & $\mathbf{93.8\%}$   \\
\hline
DS-SingleGRU & $44.97\pm 5.80$  & $90.2\%$    \\
\hline
DS-noIndvConv & $52.15 \pm 6.24$ &  $88.3\%$  \\
\hline
DS-noMergeConv & $53.06\pm 6.59$  & $87.5\%$   \\
\hline
Sensor-fusion & $606.59\pm 56.57$  &  \diagbox[dir=SW,width=3.1cm, height=0.35cm]{}{}  \\
\hline
eNav (w/o GPS) & \diagbox[dir=SW,width=2.3cm, height=0.35cm]{}{} & $6.7\%$  \\
\hline
\end{tabular}
\vspace{-0.35cm}
\end{table}

During the whole evaluation, we regard filtered GPS signal as ground truth.  CarTrack is a regression problem. Therefore, we first evaluate all algorithms with mean absolute error (MAE) between predicted and true final displacements with $95\%$ confidence interval except for the eNav (w/o GPS) algorithm, which is a map-aided algorithm without tracking real trajectories. The results about mean absolute errors are illustrated in the second column of Table~\ref{tab:carMAE}.

Compared with senior-fusion algorithm, DeepSense reduces the tracking error by an order of magnitude, which is mainly attributed to its capability to learn the composition of noise model and physical laws.
Then, we compare our DeepSense model with three variants as mentioned before. The results show the effectiveness of each designing component of our DeepSense model. The individual and merge convolutional subnets learn the interaction within and among sensor measurements respectively. The stacked recurrent structure increases the capacity of model more efficiently. Removing any component will cause performance degradation.

DeepSense model achieves $40.43\pm 5.24 m$ mean absolute error. This is almost equivalent to half of traditional city blocks ($80m\times 80m$), which means that, with the aid of map and the assumption that car is driving on roads, DeepSense model has a high probability to provide accurate trajectory tracking.
Therefore, we propose a naive map-aided track method here. For each segment of original tracking trajectory, we assign them to the most probable road segment on map (i.e., the nearest road segment on map). We then compare the resulted trajectory with ground truth. If all the trajectory segments are the same as the ground truth, we regard it as a successful tracking trajectory. Finally, we compute the percentage of successful tracking trajectories as accuracy. eNav (w/o GPS) is a map-aided algorithm, so we directly compare the trajectory segments. Sensor-fusion algorithm generates tracking errors that are comparable to driving distances, so we exclude it from the comparison.
We show the accuracy of map-aided versions of algorithms in the third column of Table~\ref{tab:carMAE}.  DeepSense outperforms eNav (w/o GPS) with a large margin, because eNav (w/o GPS) intrinsically depends on occasional on-demand GPS samples to correct tracking error.

We next examine how tracking performance is affected by driving distances.
We first sort all evaluation samples according to driving distance. Then we separate them into $10$ groups with $200$m step size. Finally, we compute mean absolute error and accuracy of map-aided track for DeepSense algorithm separately for each group.
We illustrate the results in Fig.~\ref{fig:carTrack-metricDistant}. For the mean absolute error metric, driving longer distance generally results in large error, {\em but the error does not accumulate linearly over distance}. There are mainly two reasons for this phenomenon. On one hand, we observe that the error of our predicted trajectory usually occurs during the beginning of the driving, where uncertainty in predicting driving direction is the major cause. This is also the motivation that we add the penalty term for cost function in Section~\ref{sec:mobileTaskTrain}. On the other hand, longer-driving cases in our testing samples are more stable, because we extract the trajectory from zero-speed to zero-speed. For the map-aided track, longer driving distances even yields slightly better accuracy. This is because long-distance trajectory usually contains long trajectory segments, which can help to find the ground truth on the map.

\begin{figure}[!htb]
\vspace{-0.3cm}
\begin{subfigure}{.5\linewidth}
  \centering
  \includegraphics[width=1.1\linewidth]{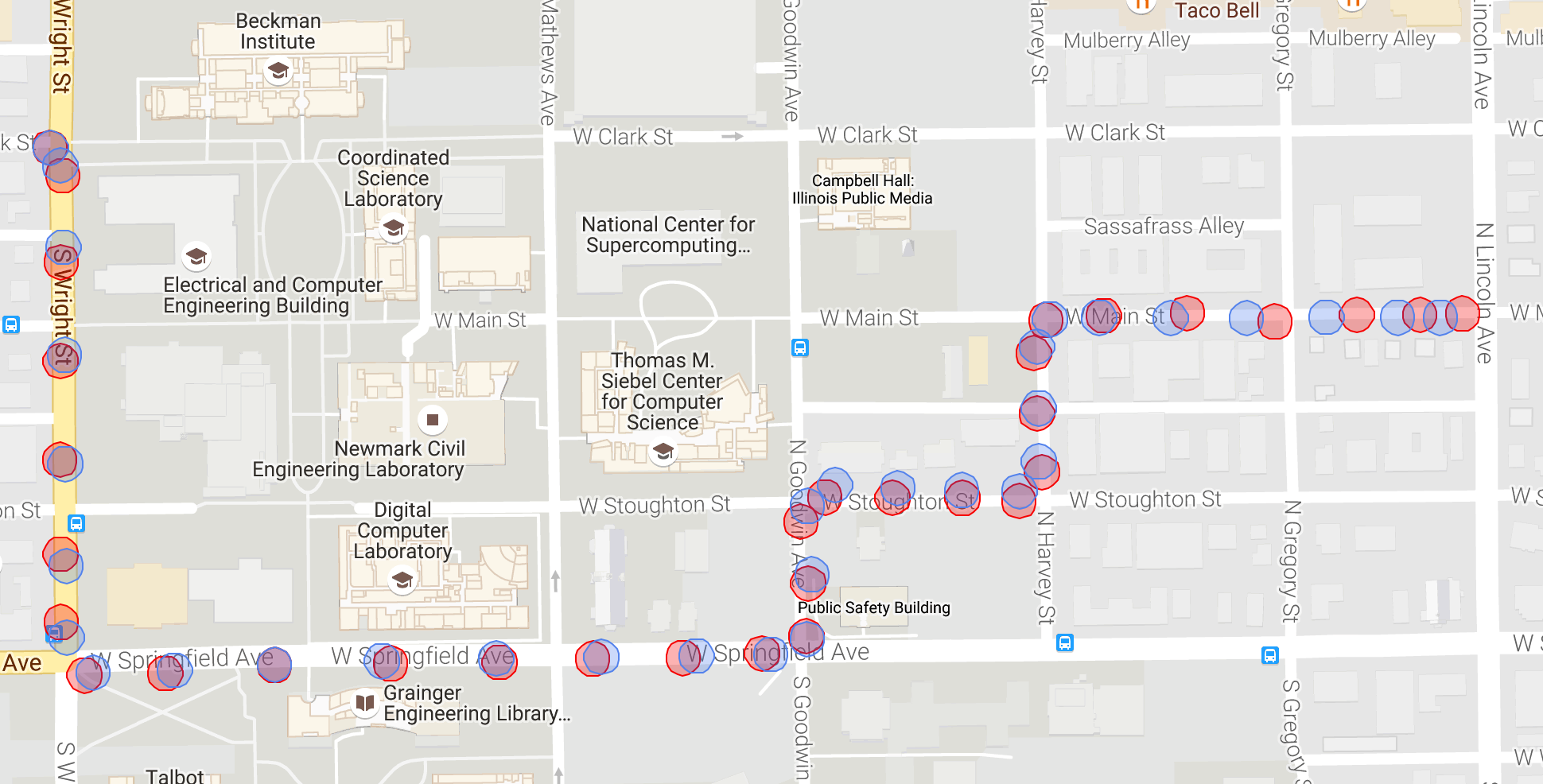}
  \caption{Trajectory a}
  \label{fig:carRes1}
\end{subfigure}%
\begin{subfigure}{.5\linewidth}
  \centering
  \includegraphics[width=.7\linewidth]{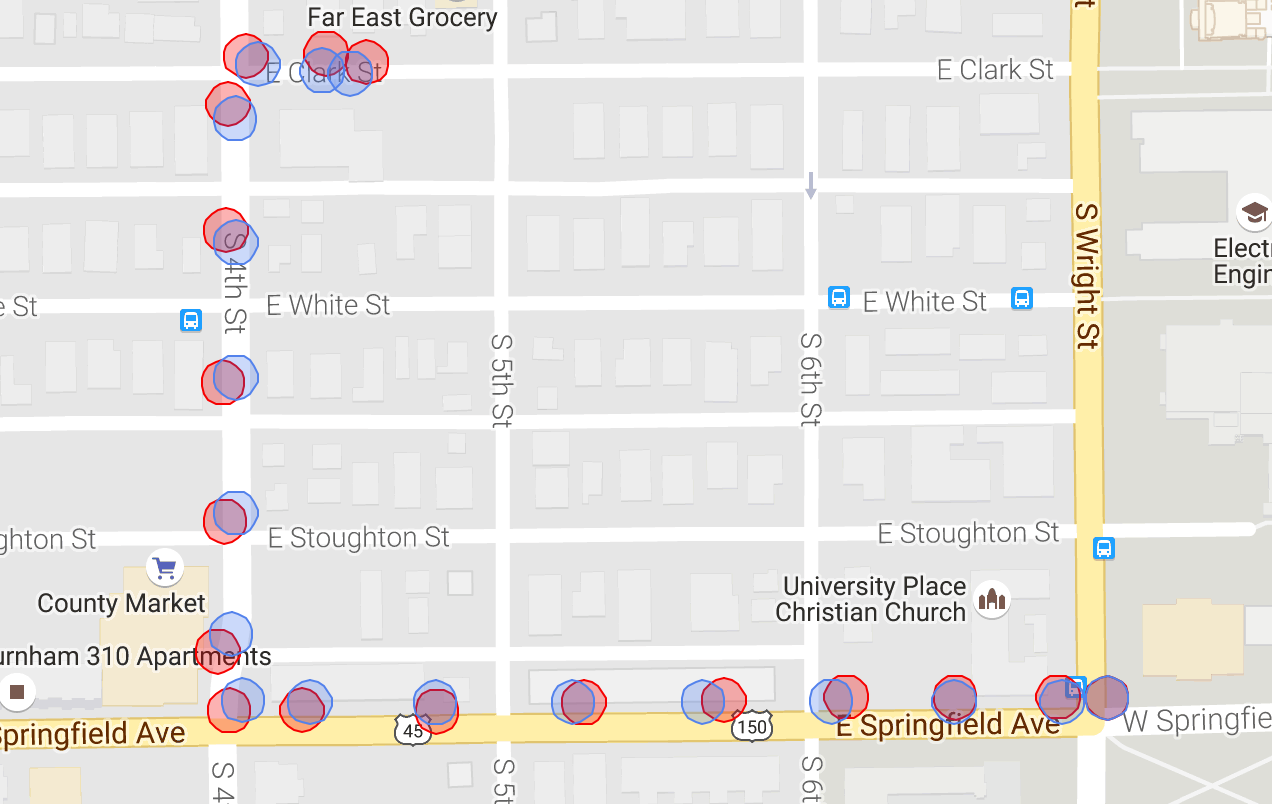}
  \caption{Trajectory b}
  \label{fig:carRes2}
\end{subfigure}
\begin{subfigure}{.5\linewidth}
  \centering
  \includegraphics[width=.7\linewidth]{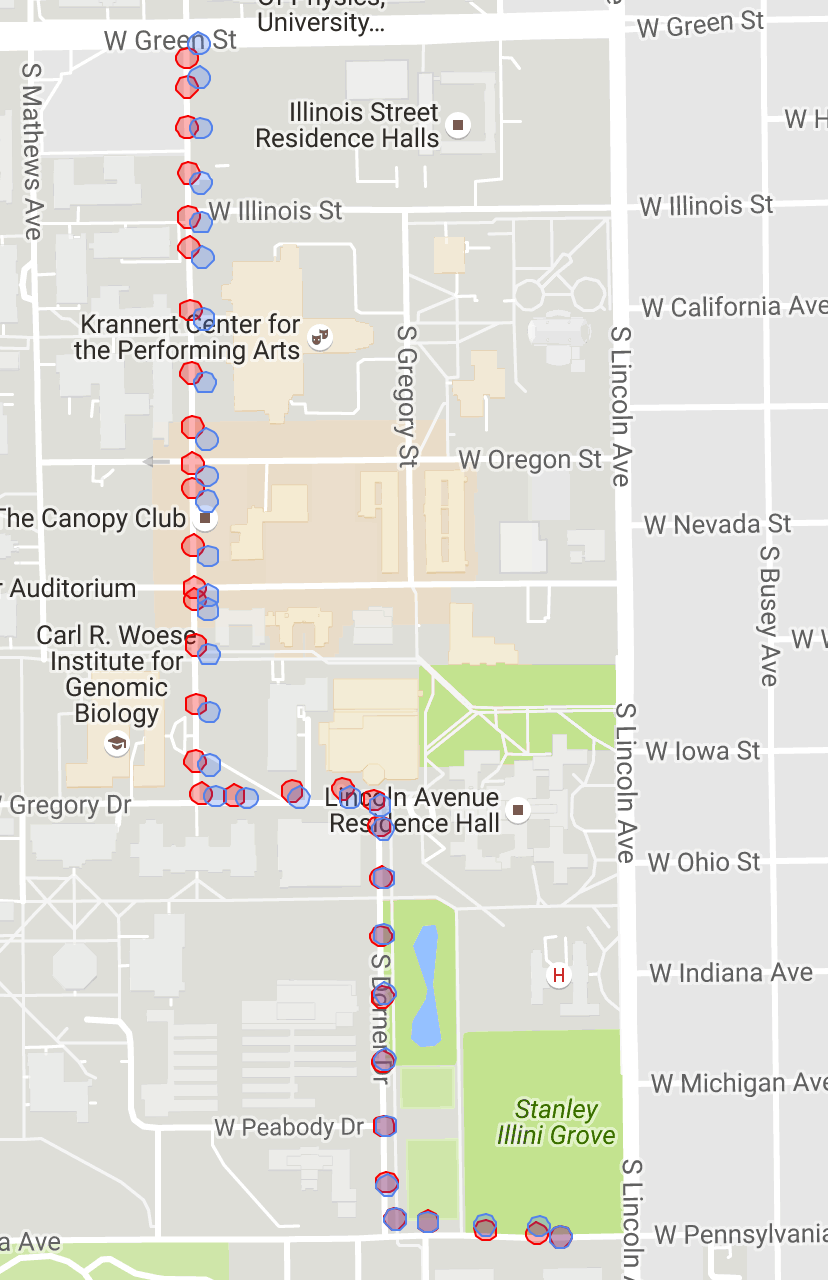}
  \caption{Trajectory c}
  \label{fig:carRes3}
\end{subfigure}
\begin{subfigure}{.5\linewidth}
  \centering
  \includegraphics[width=.7\linewidth]{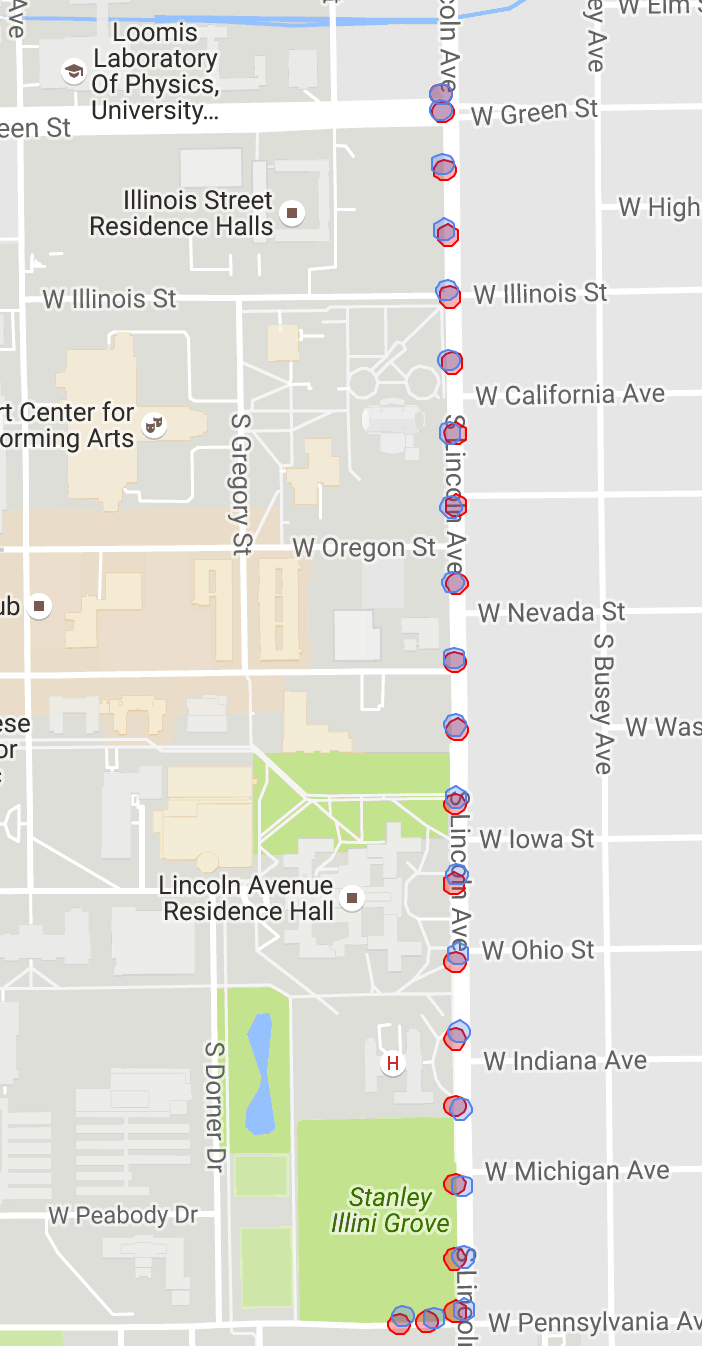}
  \caption{Trajectory d}
  \label{fig:carRes4}
\end{subfigure}
\caption{Examples of tracking trajectory without the help of map: Blue trajectory (DeepSense) and Red trajectory (GPS)}
\label{fig:carRes}
\vspace{-0.15cm}
\end{figure}

Finally, some our DeepSense tracking results (without the help of map and with downsampling) are illustrated in Fig.~\ref{fig:carRes}.

\subsubsection{HHAR}

For HHAR task, we perform leave-one-user-out evaluation (i.e., leaving the whole data from one user as testing data) on datasets consisting of 9 users, which are labelled from $a$ to $i$. We illustrate the result of evaluations according to three metrics: accuracy, macro $F_1$ score, and micro $F_1$ score with $95\%$ confidence interval in Fig.~\ref{fig:HHAR-bar}.

\begin{figure}[!htb]
\vspace{-0.3cm}
\centering
\includegraphics[width=0.75\linewidth]{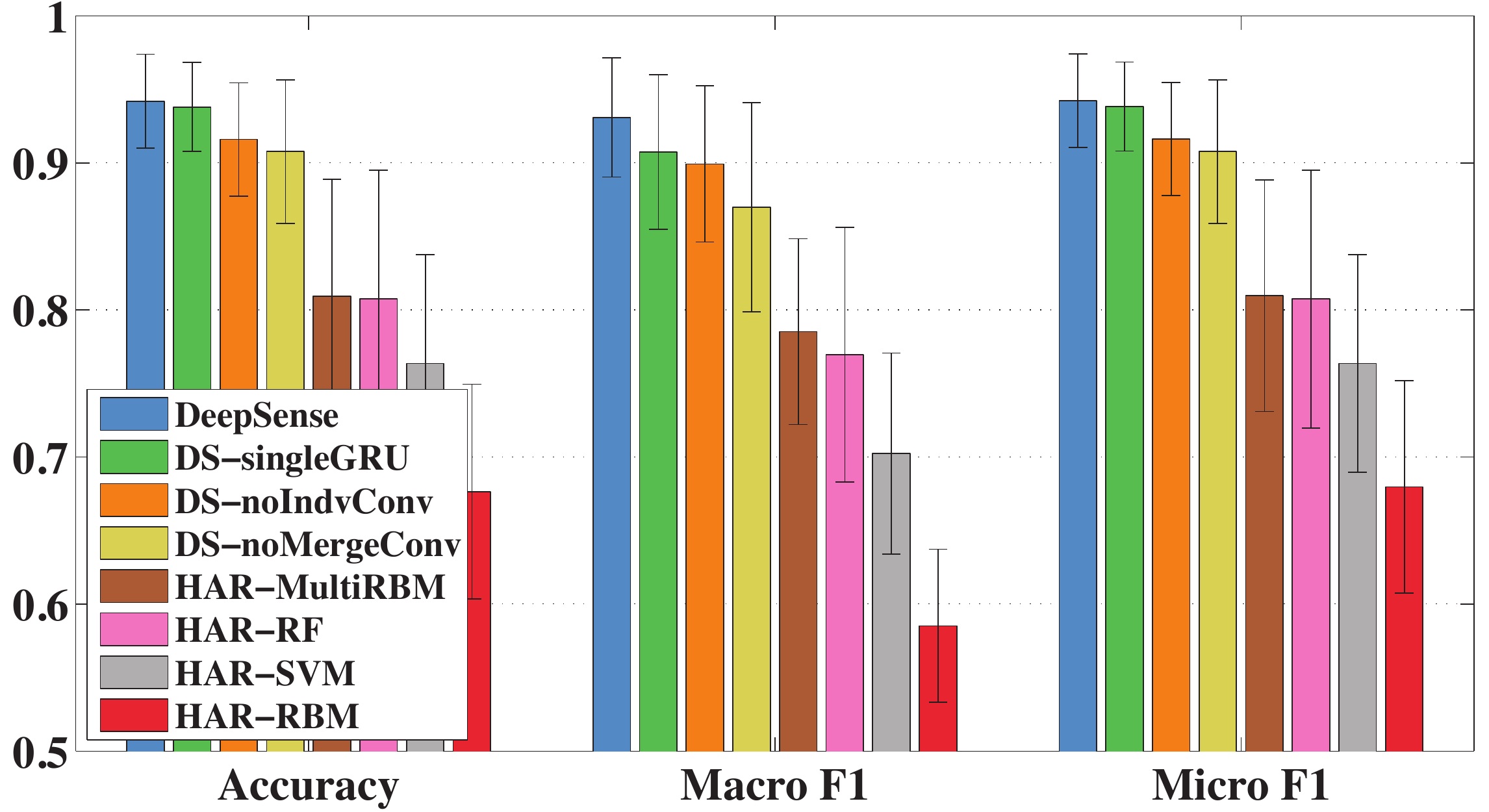}
\caption{Performance metrics of HHAR task.}
\label{fig:HHAR-bar}
\vspace{-0.05cm}
\end{figure}

\begin{figure}[!htb]
\vspace{-0.2cm}
\begin{subfigure}{.5\linewidth}
  \centering
  \includegraphics[width=1.\linewidth]{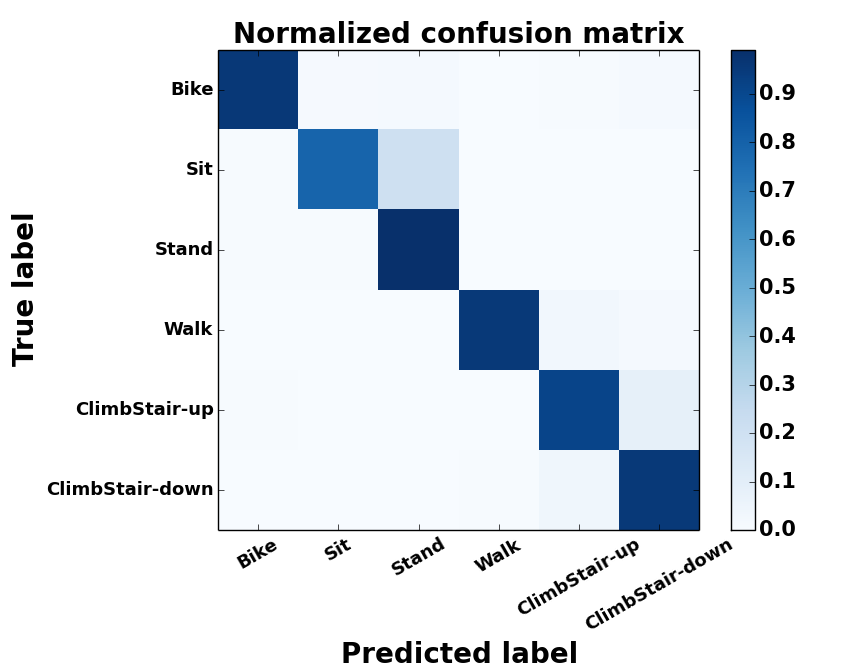}
  \caption{Confusion matrix of HHAR task.}
  \label{fig:HHAR-cm}
\end{subfigure}%
\begin{subfigure}{.5\linewidth}
  \centering
 \includegraphics[width=0.9\linewidth]{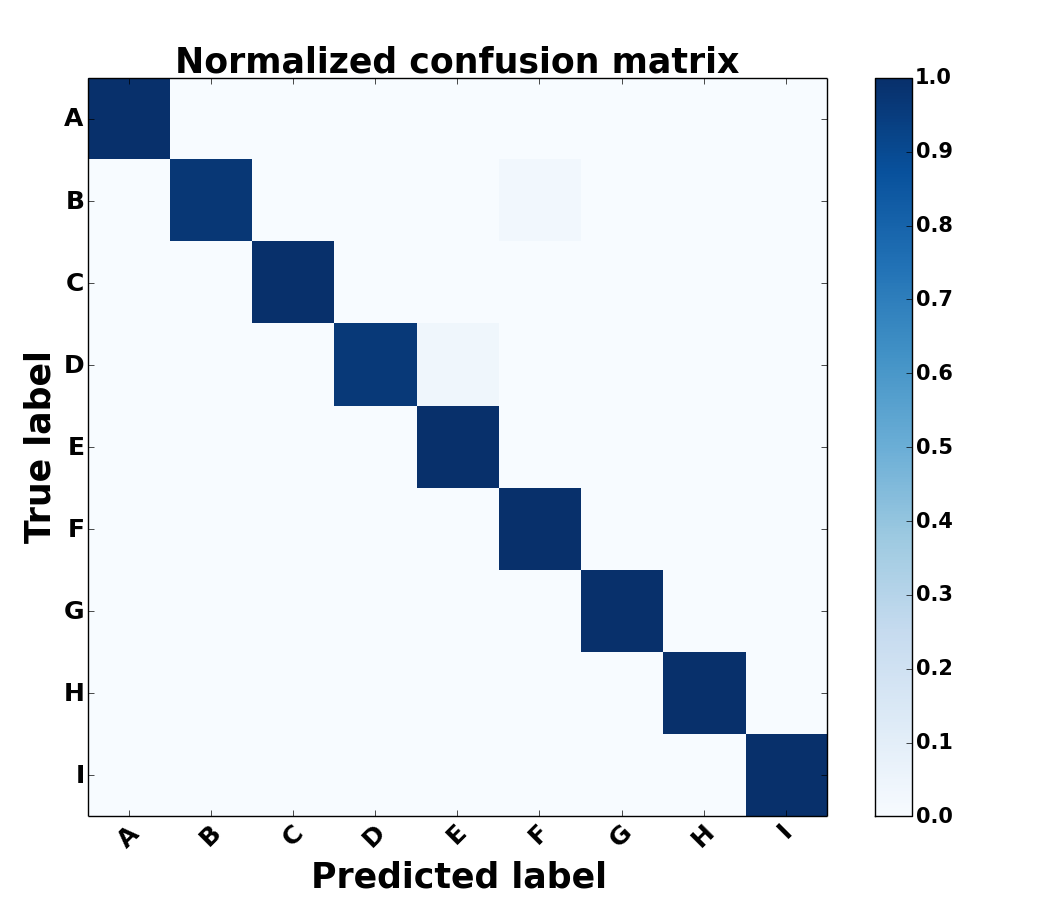}
  \caption{Confusion matrix of UserID task.}
  \label{fig:UserID-cm}
\end{subfigure}
\caption{Confusion matrix of HHAR and UserID tasks.}
\label{fig:cm}
\end{figure}

The DeepSense based algorithms (including DeepSense and three variants) outperform other baseline algorithms with a large margin (i.e., at least $10\%$). Compared with two hand-crafted feature based algorithms HAR-RF and HAR-SVM, DeepSense model can automatically extract more robust features, which generalize better to the user who does not appear in the training set. Compared with a deep model, such as HAR-RBM and HAR-MultiRBM, DeepSense model exploit local structures within sensor measurements, dependency along time, and relationships among multiple sensors to generate better and more robust features from data.
Compared with three variants, DeepSense still achieves the best performance (accuracy: $0.942\pm0.032$, macro $F_1$: $0.931\pm0.041$, and micro $F_1$: $0.942\pm0.032$). This reinforces the effectiveness of our design components in DeepSense model.

Then we illustrate the confusion matrix of best-performing DeepSense model in Fig.~\ref{fig:HHAR-cm}. Predicting $Sit$ as $Stand$ is the largest error. It is hard to classify these two, because two activities should have similar motion sensor measurements by nature, especially when we have no prior information about testing users. In addition, the algorithm has a minor error about misclassification between $ClimbStair$-$up$ and $ClimbStair$-$down$.

\subsubsection{UserID}
This task focuses on user identification with biometric motion analysis.
We evaluate all algorithms with 10-fold cross validation.
We illustrate the result of evaluations according to three metrics: accuracy, macro $F_1$ score, and micro $F_1$ score with $95\%$ confidence interval in Fig.~\ref{fig:UserID-bar}. Specifically, Fig.~\ref{fig:UserID-bar-5} shows the results when algorithms observe $1.25$ seconds of evaluation data, Fig.~\ref{fig:UserID-bar-20} shows the results when algorithms observe $5$ seconds of evaluation data.

\begin{figure}[!htb]
\vspace{-0.cm}
\begin{subfigure}{.5\linewidth}
  \centering
  \includegraphics[width=1.\linewidth]{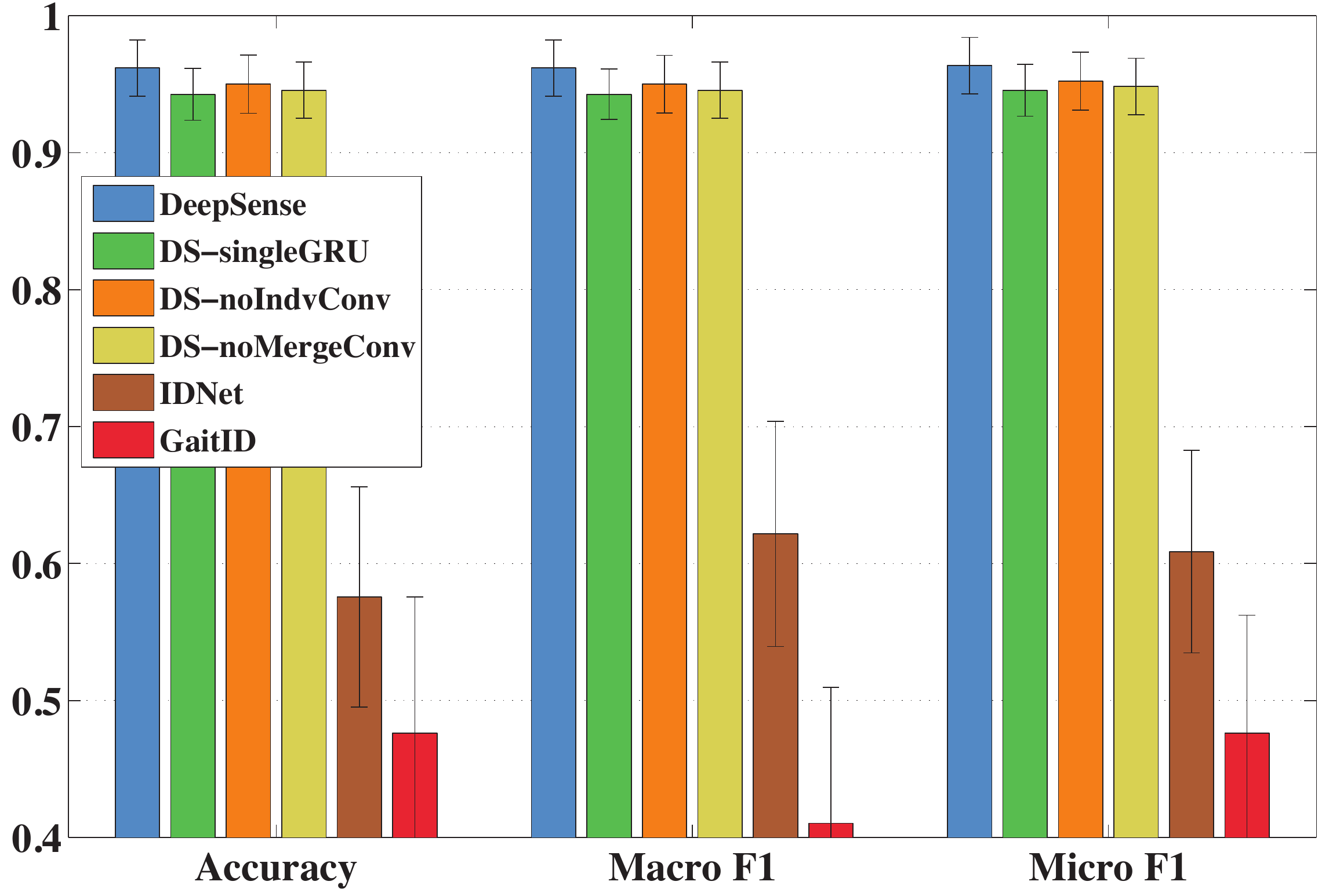}
  \caption{$5$ time intervals: $1.25$s}
  \label{fig:UserID-bar-5}
\end{subfigure}
\begin{subfigure}{.5\linewidth}
  \centering
 \includegraphics[width=1.\linewidth]{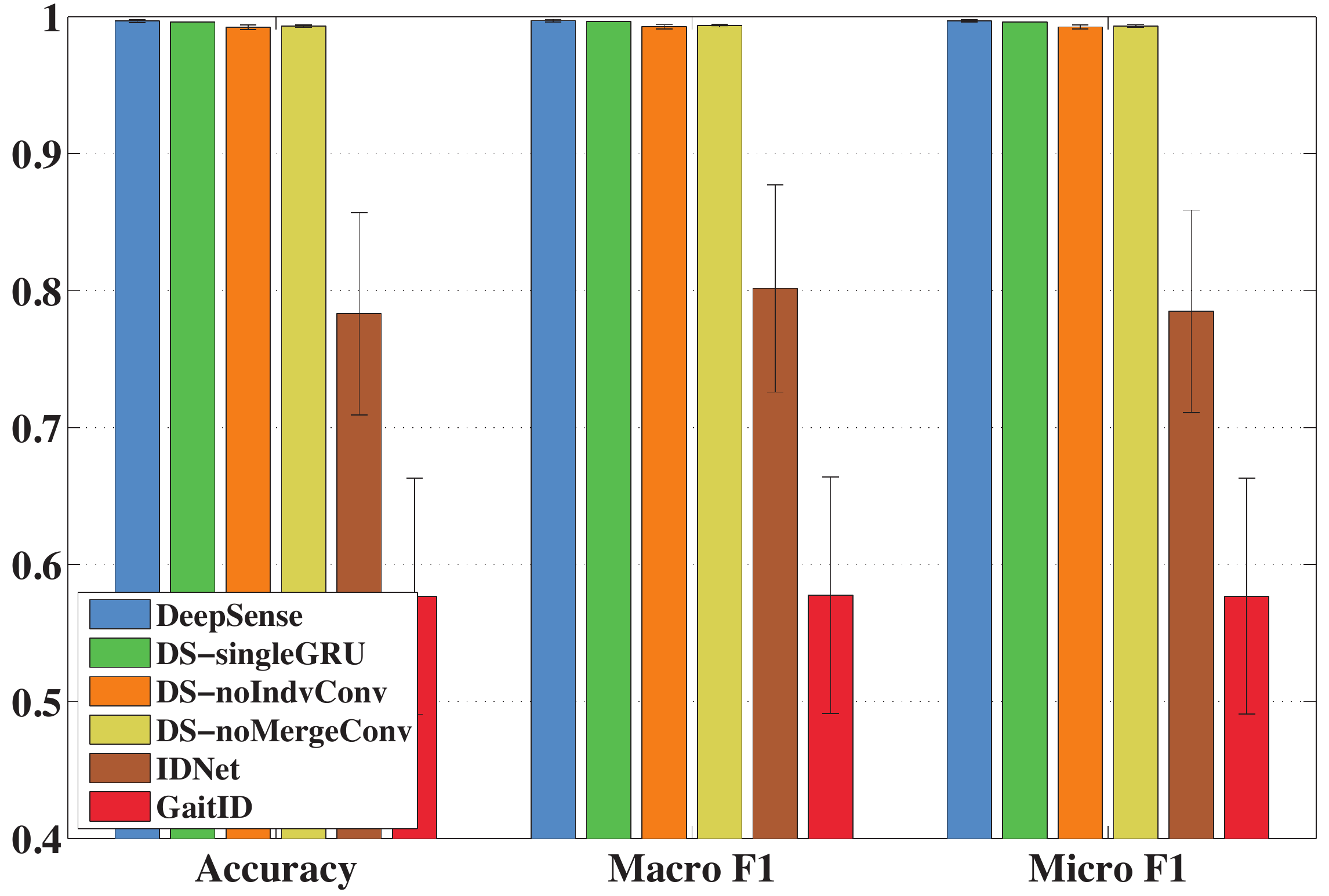}
  \caption{$20$ time intervals: $5$s}
  \label{fig:UserID-bar-20}
\end{subfigure}
\caption{Performance metrics of UserID task.}
\label{fig:UserID-bar}
\end{figure}

\begin{figure}[!htb]
\vspace{-0.15cm}
\centering
\includegraphics[width=0.75\linewidth]{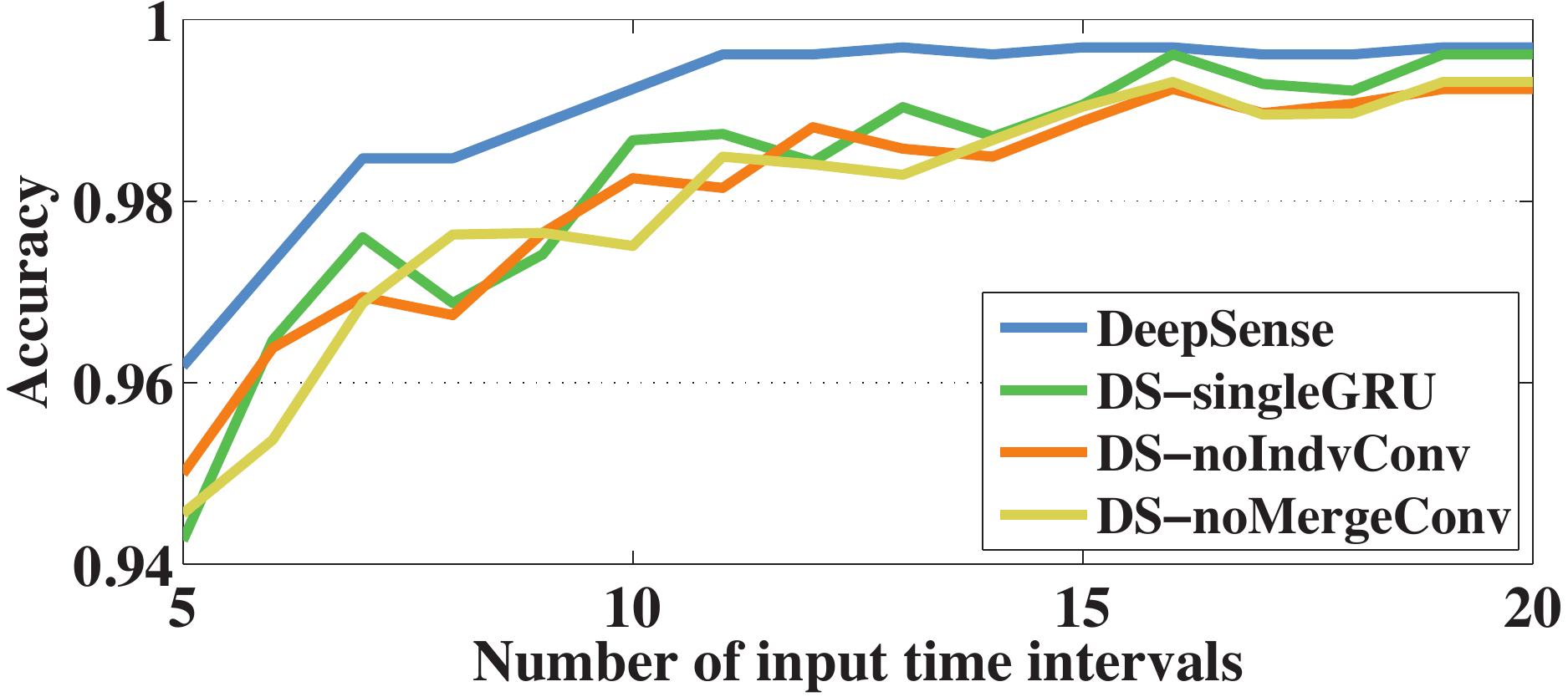}
\caption{Accuracy over input measurement length of UserID task.}
\label{fig:UserID-step}
\vspace{-0.cm}
\end{figure}

DeepSense and three variants outperform other baseline algorithms with a large margin again (i.e. at least 20\%). Compared with the template extraction and matching method, GaitID, DeepSense model can automatically extract distinct features from data, which fit well to not only walking but also all other kinds of activities. Compared with method that first extracts templates and then apply neural network to learn features, IDNet, DeepSense solves the whole task in the end-to-end fashion. We eliminate the manually processing part and exploit local, global, and temporal relationships through our architecture, which results better performance. In this task, although the performance of different variants is similar when observing data with $5$ seconds, DeepSense still achieves the best performance (accuracy: $0.997\pm 0.001$, macro $F_1$: $0.997\pm0.001$, and micro $F_1$: $0.997\pm 0.001$).

We further compare DeepSense with three variants by changing the number of evaluation time intervals from $5$ to $20$, which corresponds to around $1$ to $5$ seconds. We compute the accuracy for each case. The results illustrated in Fig.~\ref{fig:UserID-step} suggest that DeepSense performs better than all the other variants with a relatively large margin when algorithms observe sensing data with shorter time.
This indicates the effectiveness of design components in DeepSense.

Then we illustrate the confusion matrix of best-performing DeepSense model when observing sensing data with $5$ seconds in Fig.~\ref{fig:UserID-cm}. It shows
that the algorithm gives
a pretty good result. On average, only about two misclassifications appear during each testing.

\subsection{Latency and Energy}

Final, we examine the computation latency and energy consumption of DeepSense---stereotypical deep learning models are traditionally power hungry and time consuming---we illustrate, through our careful measurements in all three example application scenarios, the feasibility of directly implementing and deploying DeepSense on mobile devices without any additional optimization.

\begin{figure}[!htb]
\vspace{-0.3cm}
\centering
\includegraphics[width=0.6\linewidth]{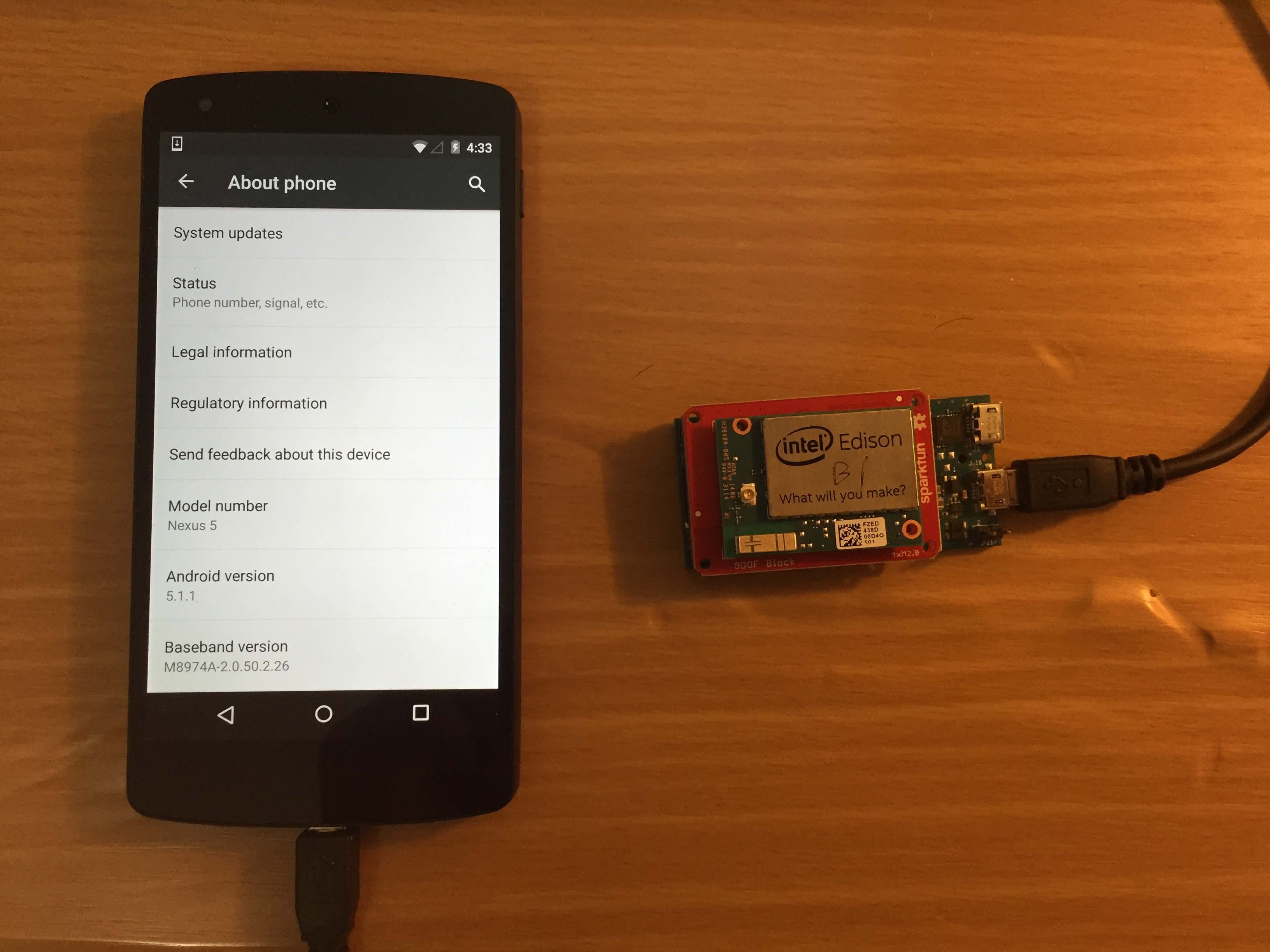}
\caption{Test Platforms: Nexus5 and Intel Edison.}
\label{fig:testPlatforms}
\vspace{-0.05cm}
\end{figure}

\begin{figure}[!htb]
\begin{subfigure}{.5\linewidth}
  \centering
  \includegraphics[width=1.\linewidth]{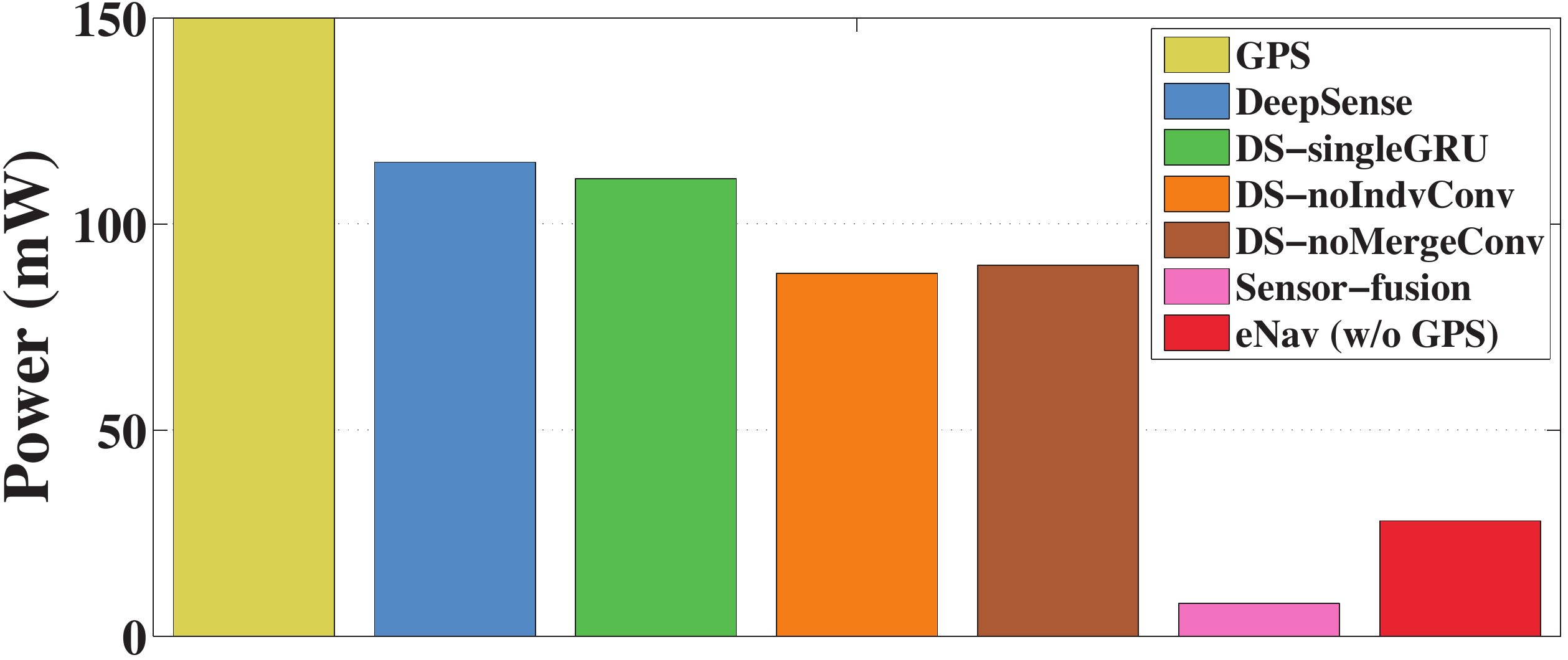}
  \caption{Power}
  \label{fig:carTrackEnergy}
\end{subfigure}%
\begin{subfigure}{.5\linewidth}
  \centering
  \includegraphics[width=1.\linewidth]{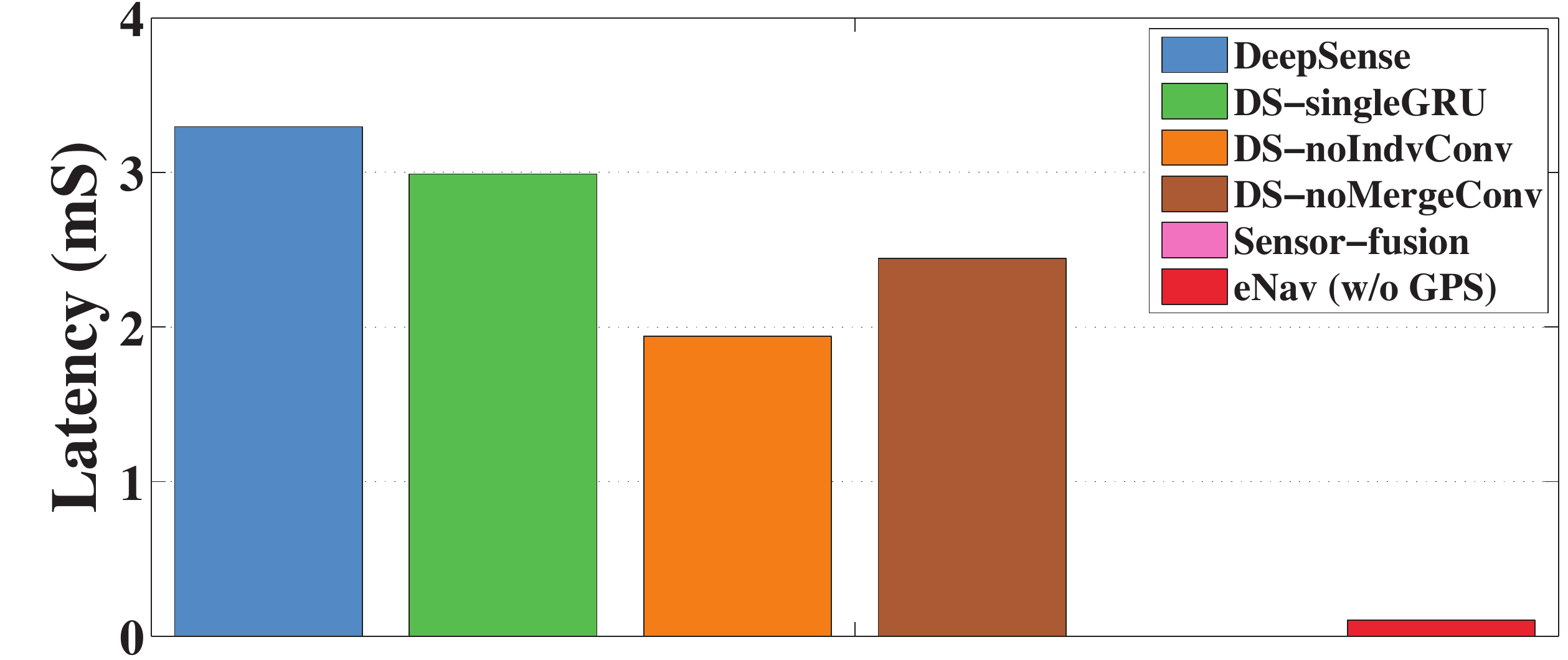}
  \caption{Latency}
  \label{fig:carTrackLatency}
\end{subfigure}
\caption{Power and Latency of carTrack solutions on Nexus 5}
\label{fig:carTrackEnergyTime}
\vspace{-0.2cm}
\end{figure}

\begin{figure}[!htb]
\begin{subfigure}{.5\linewidth}
  \centering
  \includegraphics[width=1.\linewidth]{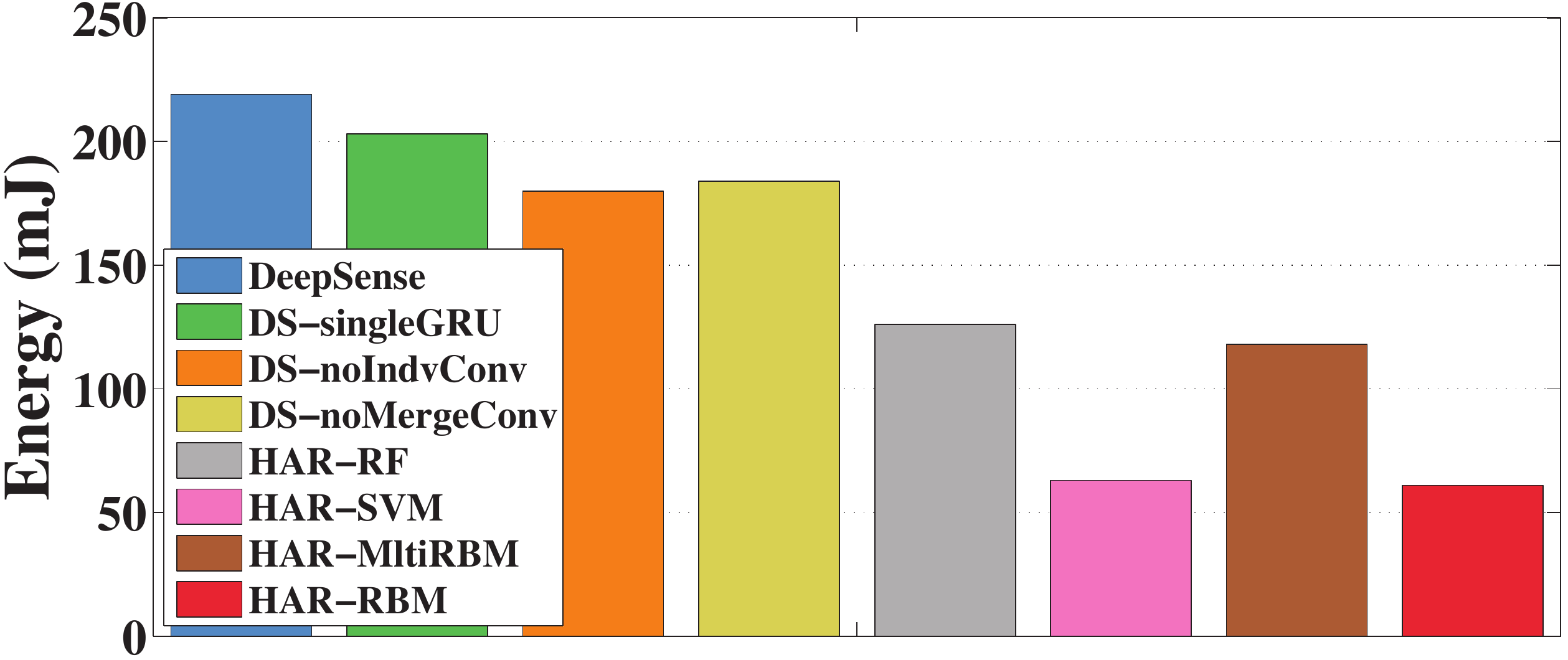}
  \caption{Energy}
  \label{fig:HHAREnergy}
\end{subfigure}%
\begin{subfigure}{.5\linewidth}
  \centering
  \includegraphics[width=1.\linewidth]{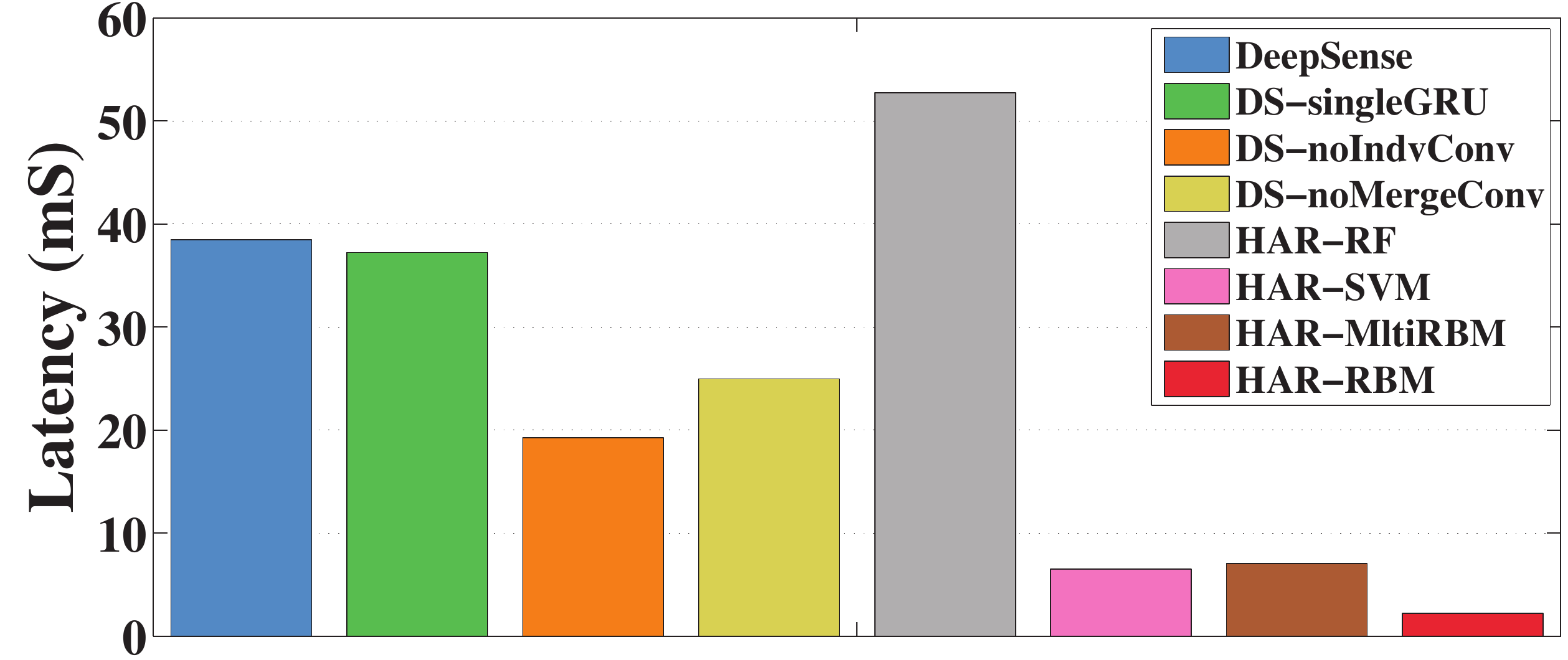}
  \caption{Latency}
  \label{fig:HHARLatency}
\end{subfigure}
\caption{Energy and Latency of HHAR solutions on Nexus 5}
\label{fig:HHAREnergyTime}
\vspace{-0.2cm}
\end{figure}

\begin{figure}[!htb]
\begin{subfigure}{.5\linewidth}
  \centering
  \includegraphics[width=1.\linewidth]{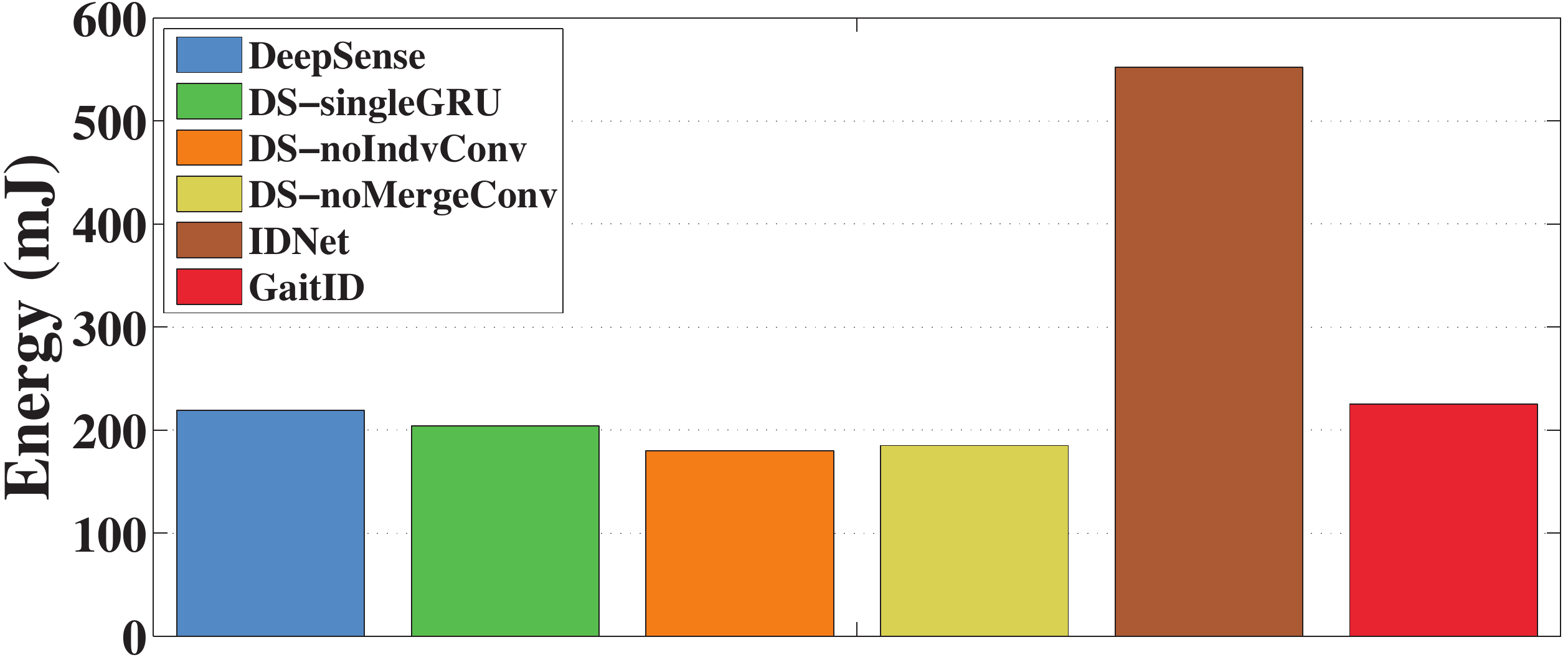}
  \caption{Energy}
  \label{fig:UserIDEnergy}
\end{subfigure}%
\begin{subfigure}{.5\linewidth}
  \centering
  \includegraphics[width=1.\linewidth]{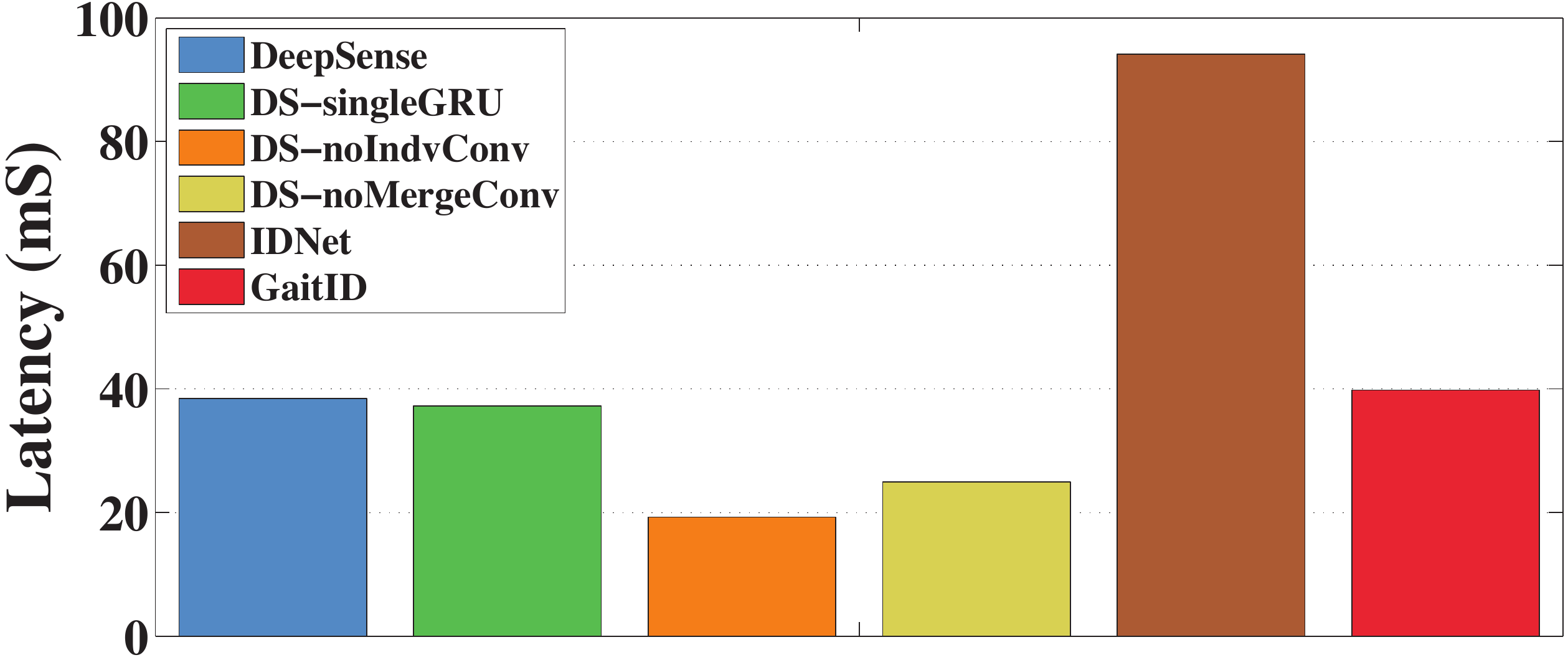}
  \caption{Latency}
  \label{fig:UserIDLatency}
\end{subfigure}
\caption{Energy and Latency of UserID solutions on Nexus 5}
\label{fig:UserIDEnergyTime}
\vspace{-0.2cm}
\end{figure}

\begin{figure}[!htb]
\begin{subfigure}{.5\linewidth}
  \centering
  \includegraphics[width=1.\linewidth]{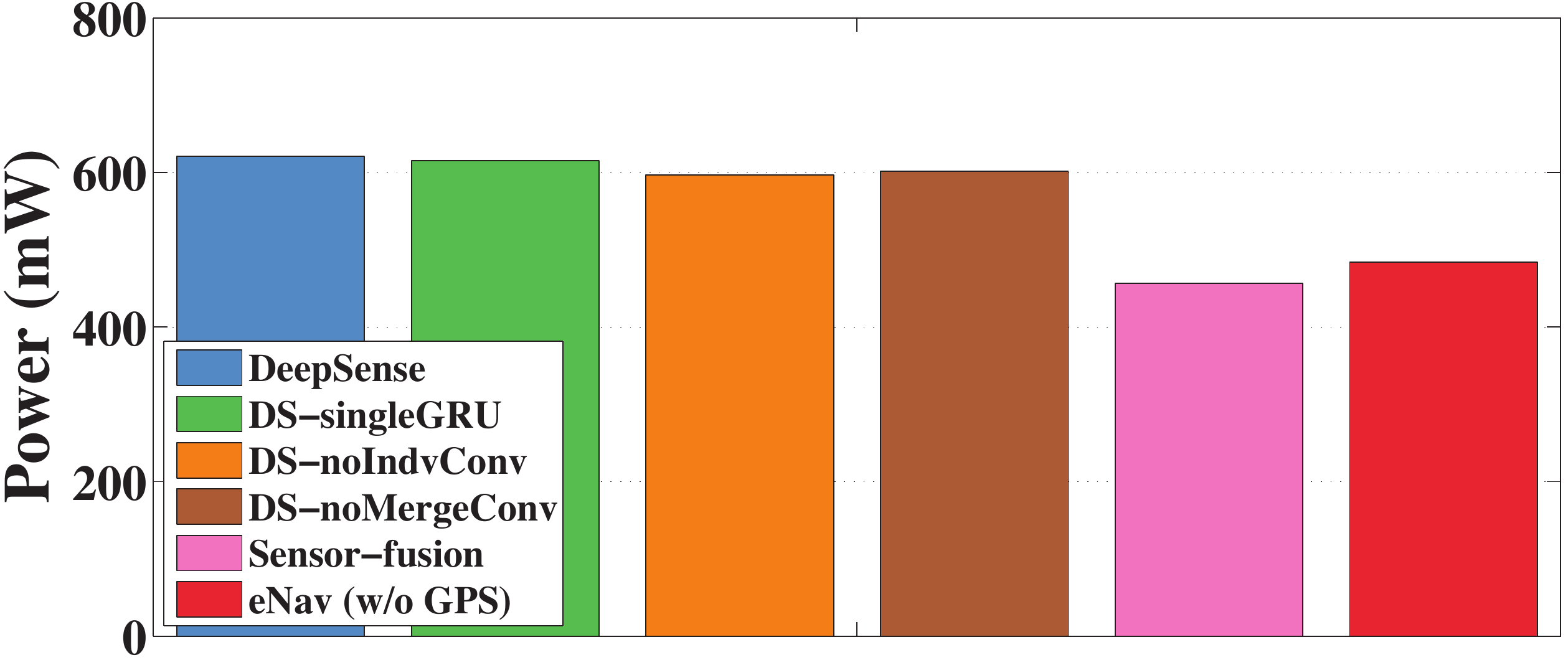}
  \caption{Power}
  \label{fig:carTrackEnergyEdison}
\end{subfigure}%
\begin{subfigure}{.5\linewidth}
  \centering
  \includegraphics[width=1.\linewidth]{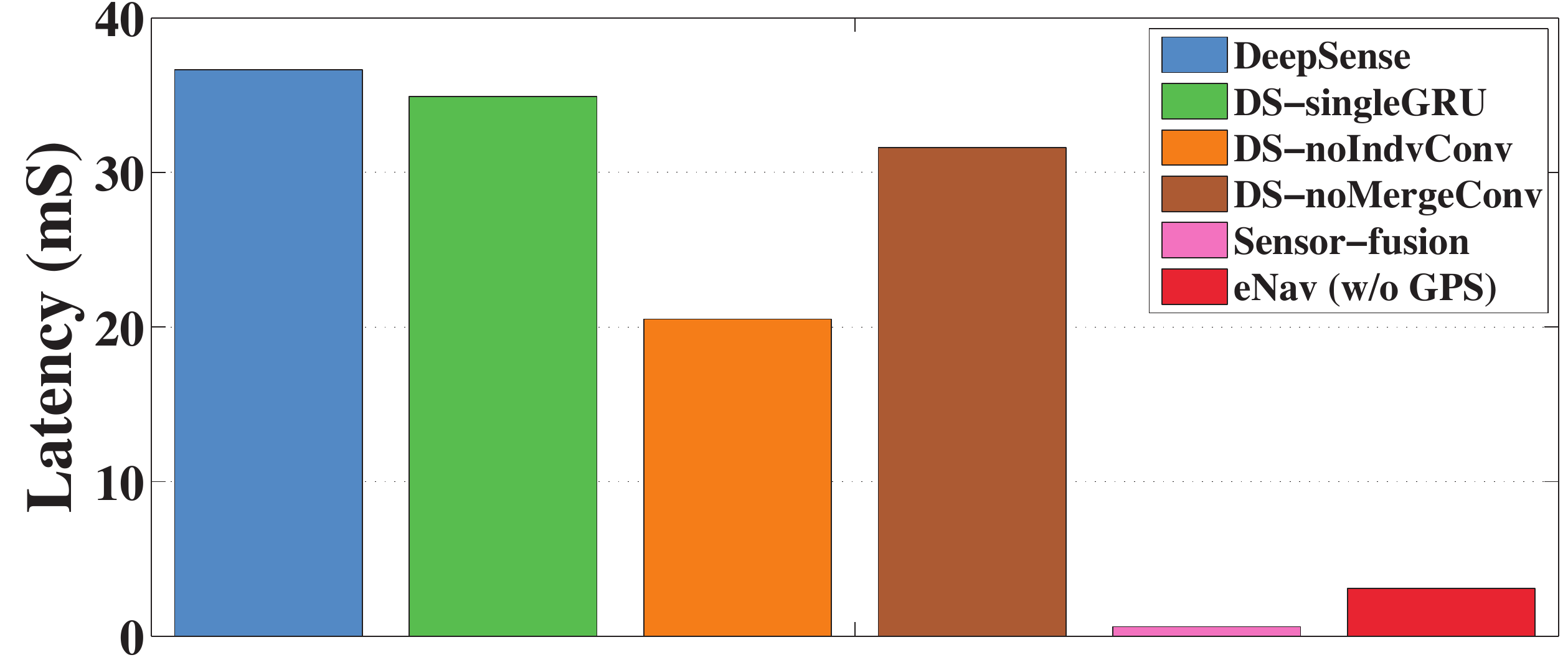}
  \caption{Latency}
  \label{fig:carTrackLatencyEdison}
\end{subfigure}
\caption{Power and Latency of carTrack solutions on Intel Edison}
\label{fig:carTrackEnergyTimeEdison}
\vspace{-0.2cm}
\end{figure}

\begin{figure}[!htb]
\begin{subfigure}{.5\linewidth}
  \centering
  \includegraphics[width=1.\linewidth]{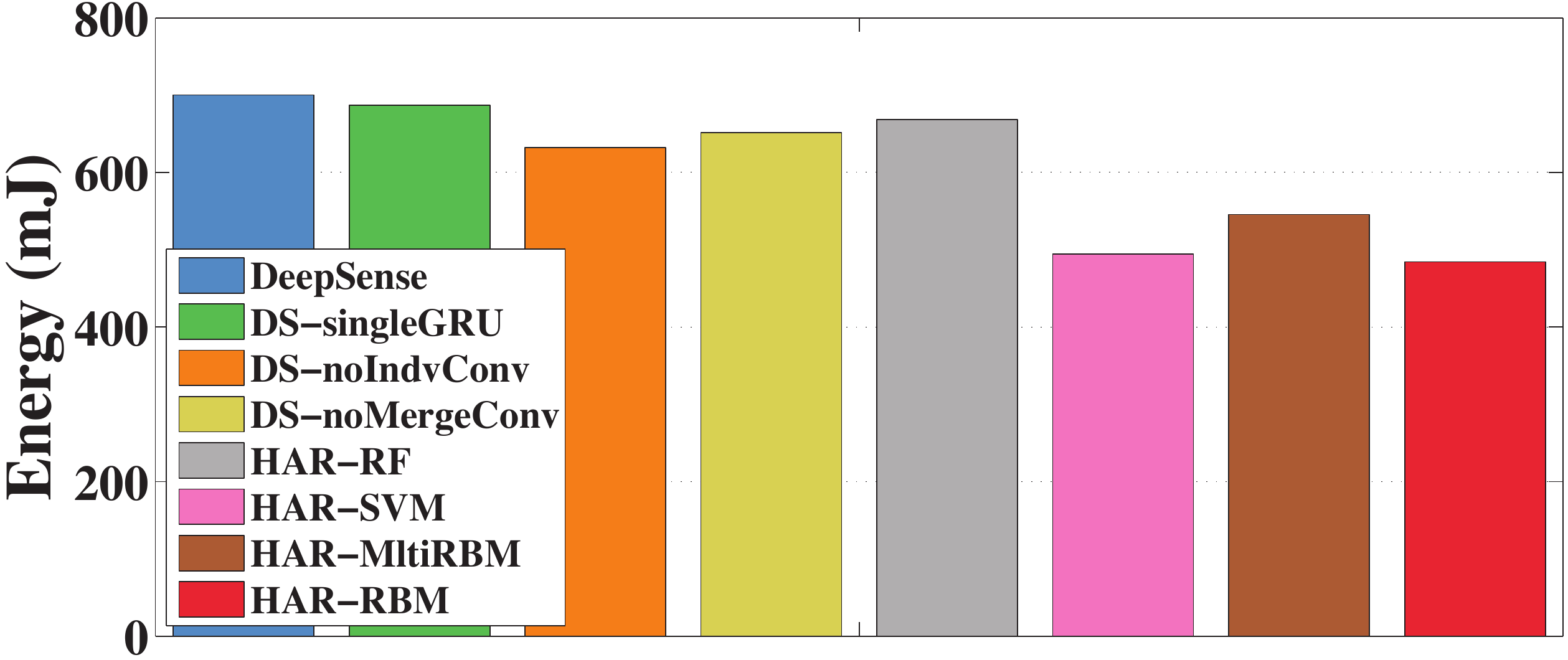}
  \caption{Energy}
  \label{fig:HHAREnergyEdison}
\end{subfigure}%
\begin{subfigure}{.5\linewidth}
  \centering
  \includegraphics[width=1.\linewidth]{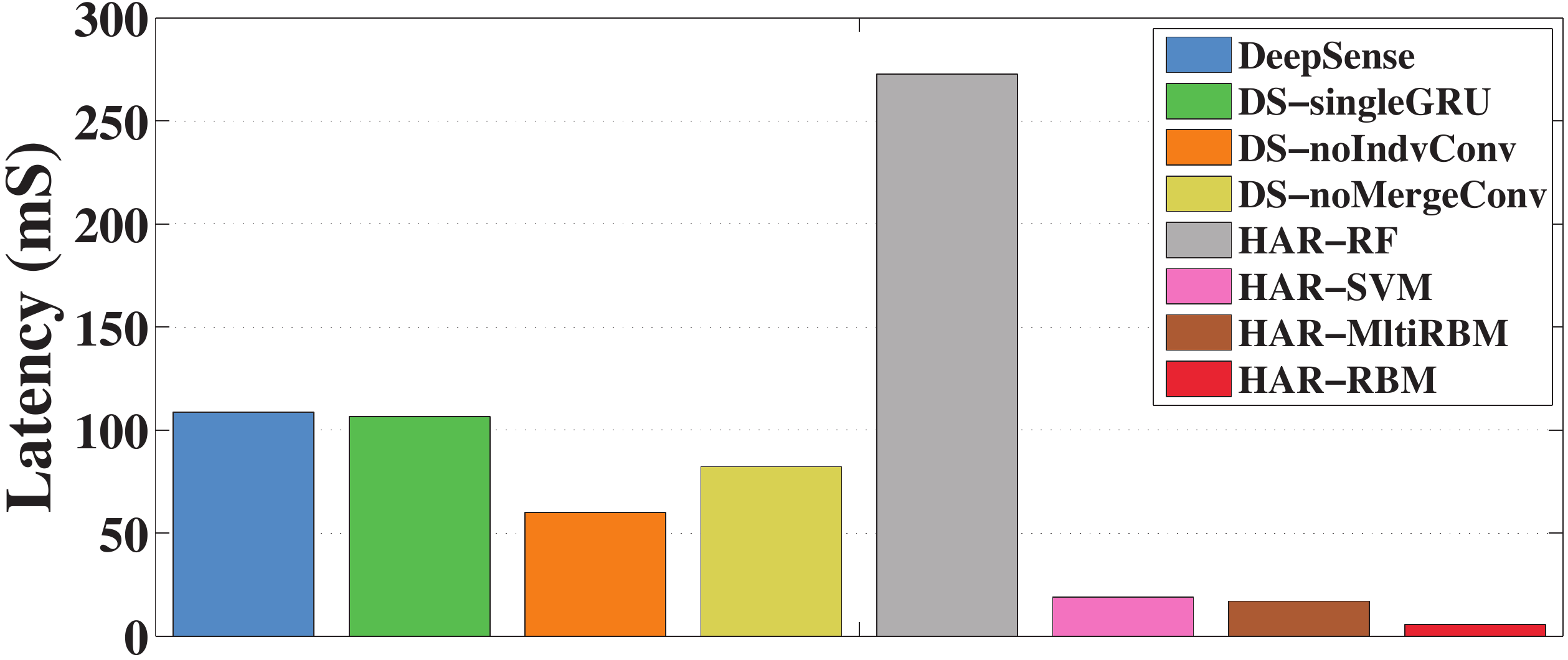}
  \caption{Latency}
  \label{fig:HHARLatencyEdison}
\end{subfigure}
\caption{Energy and Latency of HHAR solutions on Intel Edison}
\label{fig:HHAREnergyTimeEdison}
\vspace{-0.2cm}
\end{figure}

\begin{figure}[!htb]
\begin{subfigure}{.5\linewidth}
  \centering
  \includegraphics[width=1.\linewidth]{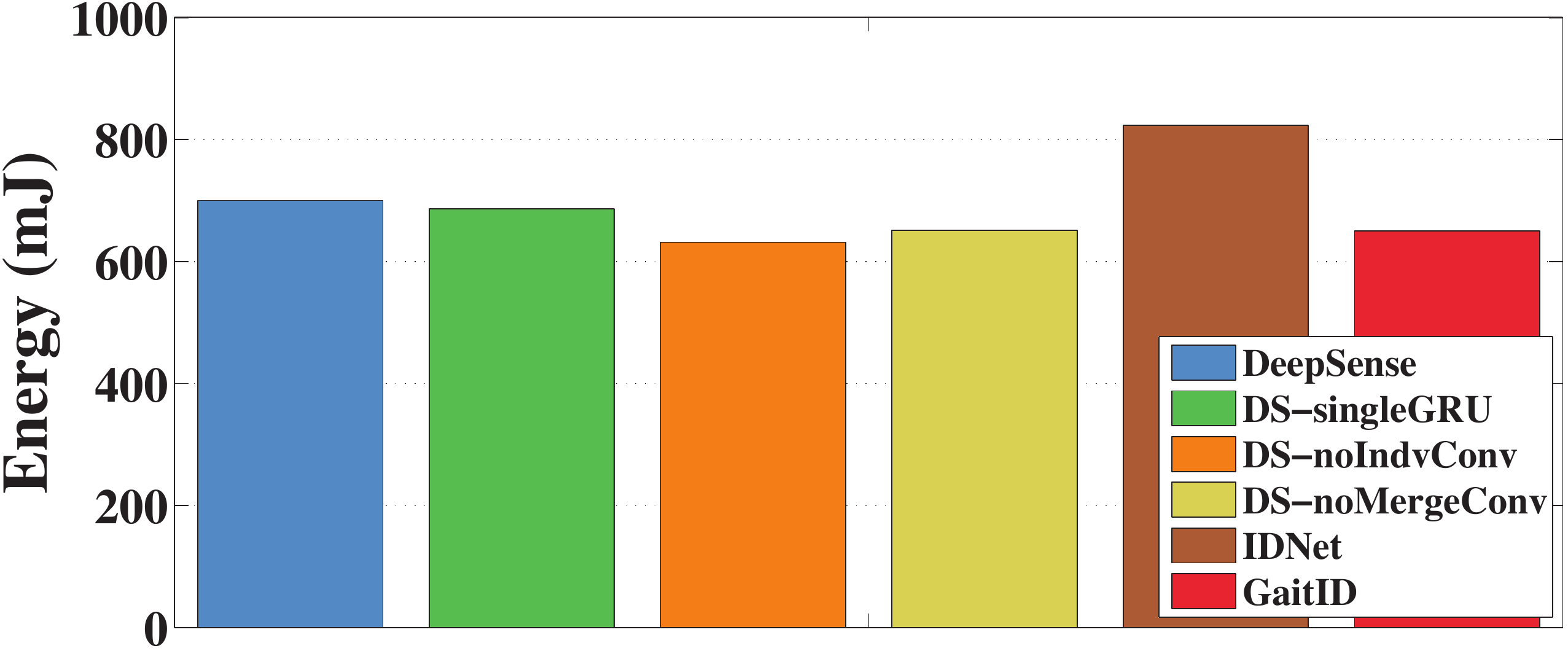}
  \caption{Energy}
  \label{fig:UserIDEnergyEdison}
\end{subfigure}%
\begin{subfigure}{.5\linewidth}
  \centering
  \includegraphics[width=1.\linewidth]{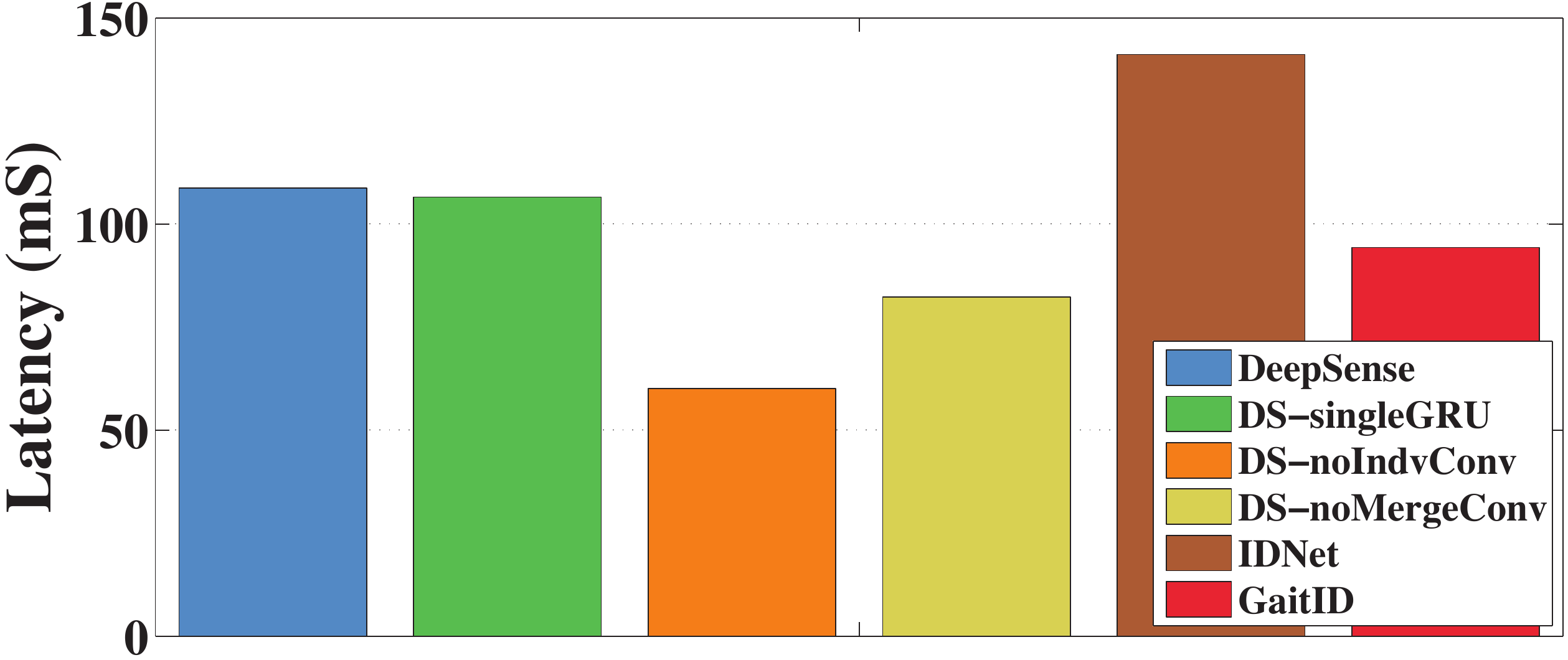}
  \caption{Latency}
  \label{fig:UserIDLatencyEdison}
\end{subfigure}
\caption{Energy and Latency of UserID solutions on Intel Edison}
\label{fig:UserIDEnergyTimeEdison}
\end{figure}

Experiments are conducted on two kinds of devices: Nexus 5 and Intel Edison, as shown in Fig.~\ref{fig:testPlatforms}. The energy consumption of applications on Nexus 5 is measured by PowerTutor~\cite{zhang2010accurate}, while the energy consumption of Intel Edison is measured by an external power monitor. The evaluations of energy and latency on Nexus 5 are shown in Fig.~\ref{fig:carTrackEnergyTime}, \ref{fig:HHAREnergyTime}, and \ref{fig:UserIDEnergyTime},
and Intel Edison Fig.~\ref{fig:carTrackEnergyTimeEdison}, \ref{fig:HHAREnergyTimeEdison}, and \ref{fig:UserIDEnergyTimeEdison}. Since algorithms for carTrack are designed to report position every second, we show the power consumption in Fig.~\ref{fig:carTrackEnergy} and \ref{fig:carTrackEnergyEdison}. Other two tasks are not periodical tasks by nature. Therefore, we show the per-inference energy consumption in Fig.~\ref{fig:HHAREnergy}, \ref{fig:HHAREnergyEdison}, \ref{fig:UserIDEnergy}, and \ref{fig:UserIDEnergyEdison}. For experiments on Intel Edison, notice that we measured total energy consumption, containing $419$mW idle-mode power consumption.

For the carTrack task, all DeepSense based models consume a bit less energy compared with 1-Hz GPS samplings on Nexus 5. The running times are measured in the order of microsecond on both platforms, which meets the requirement of per-second measurement.

For the HHAR task, all DeepSense based models take moderate energy and low latency to obtain one classification prediction on two platforms. An interesting observation is that HHAR-RF, a random forest model, has a relatively longer latency. This is due to the fact that random forest is an ensemble method, which involves combining a bag of individual decision tree classifiers.

For the UserID task, except for the IDNet baseline, all other algorithms show similar running time and energy consumption on two platforms. IDNet contains both a multi-stage pre-processing process and a relative large CNN, which takes longer time and more energy to compute in total.

\vspace{-0.1cm}
\section{Discussion}~\label{sec:discussion}
This paper focuses on solving different mobile sensing and computing tasks in a unified framework. DeepSense is our solution. It is a framework that requires only a few steps to be customized into particular tasks.  During the customization steps, we do not tailor the architecture for different tasks in order to lessen the requirement of human efforts while using the framework. However, particular changes to the architecture can bring additional performance gains to specific tasks.

One possible change is separating noise model and physical laws for regression-oriented tasks. The original DeepSense directly learns the composition of noise model and physical laws, providing the capability of automatically understanding underlying physical process from data. However, if we know exactly the physical process, we can use DeepSense as a powerful denoising component, and apply physical laws to the outputs of DeepSense.

The other possible change is removing some design components to trade accuracy for energy. In our evaluations, we show that some variants take acceptable degradation on accuracy with less energy consumption. The basic principle of removing design components is based on their functionalities. Individual convolutional subnets explore relationship within each sensor; merge convolutional subnet explores relationship among different sensors; and stacked RNN increases the model capacity for exploring relationship over time. We can choose to omit some components according to the demands of particular tasks.

In addition, although our three evaluation tasks focus mainly on motion sensors, which are the most widely deployed sensors, we can directly apply DeepSense to almost all other sensors, such as microphone, Wi-Fi signals, Barometer, and light sensor. We need further study on applying DeepSense to explore new applications on smart devices.

At last, for a particular sensing task, if there is drastic change in the physical environment, DeepSense might need to be re-trained with new data. However, on one hand, the traditional solution with pre-defined noise model and physical laws (or hand-crafted features) would also need redesigns anyways. On the other hand, an existing trained DeepSense framework can serve as a good initialization stage for the new training process that aids in optimization and reduce generalization error~\cite{dahl2012context}.

\vspace{-0.1cm}
\section{Conclusion}~\label{sec:conclusion}
In this paper we introduced our unified DeepSense framework for mobile sensing and computing tasks.
DeepSense integrates convolutional and recurrent neural networks to exploit different types of relationships in sensor inputs, thanks to which, it is able to learn the composition of physical laws and noise model for regression-oriented problems, and automatically extract robust and distinct features on local, global, and temporal domains to effectively carry out classification tasks---the two major focuses in mobile sensing literature. We evaluated DeepSense via three representative mobile sensing tasks, where DeepSense outperformed state of the art baselines by significant margins while still claiming its mobile-feasibility through moderate energy consumption and low latency on both mobile and embedded platforms. Our experience with the multiple DeepSense variants also provided us with valuable insights and promising guidelines in the opportunities of further framework adaptation and customization for a wide range of applications.

}

\begin{spacing}{0.86}
{
\newpage
\setlength{\columnsep}{0.1in}
\bibliographystyle{abbrv}
\bibliography{reference}
}
\end{spacing}

\end{document}